\documentclass[10pt,twocolumn,letterpaper]{article}
\usepackage{cvpr}
\usepackage{times}
\usepackage{epsfig}
\usepackage{amsmath}
\usepackage{amssymb}
\usepackage{authblk}
\usepackage{booktabs}
\usepackage{overpic}
\usepackage{graphicx}
\usepackage{color}
\usepackage{xcolor}
\usepackage{outlines}
\usepackage{subcaption}
\usepackage{enumitem}
\usepackage{hyperref}

\cvprfinalcopy 

\def\cvprPaperID{****} 

\ifcvprfinal\fi
\begin{document}

\title{NTIRE 2020 Challenge on Image Demoireing: Methods and Results}

\author{
\normalsize{
Shanxin Yuan \hspace*{6mm}
Radu Timofte \hspace*{6mm}
Ale\v{s} Leonardis  \hspace*{6mm}
Gregory Slabaugh \\
Xiaotong Luo \hspace*{3mm}
Jiangtao Zhang \hspace*{3mm} 
Yanyun Qu \hspace*{3mm}
Ming Hong \hspace*{3mm}
Yuan Xie \hspace*{3mm}
Cuihua Li \hspace*{3mm}
Dejia Xu \hspace*{3mm}
Yihao Chu \\
Qingyan Sun \hspace*{3mm}
Shuai Liu \hspace*{3mm} 
Ziyao Zong \hspace*{3mm} 
Nan Nan \hspace*{3mm}
Chenghua Li \hspace{3mm}
Sangmin Kim \hspace{3mm}
Hyungjoon Nam \\
Jisu Kim \hspace*{3mm} 
Jechang Jeong \hspace*{3mm}
Manri Cheon \hspace*{3mm}
Sung-Jun Yoon \hspace*{3mm}
Byungyeon Kang \hspace{3mm}
Junwoo Lee  \hspace{3mm}
Bolun Zheng \\
Xiaohong Liu \hspace*{3mm}
Linhui Dai\hspace*{3mm}
Jun Chen\hspace*{3mm}
Xi Cheng \hspace*{3mm}
Zhenyong Fu \hspace*{3mm}
Jian Yang \hspace*{3mm}
Chul Lee \hspace*{3mm}
An Gia Vien \\
Hyunkook Park \hspace*{3mm}
Sabari Nathan \hspace*{3mm}
M.Parisa Beham \hspace*{3mm}
S Mohamed Mansoor Roomi\hspace*{4mm}
Florian Lemarchand \\
Maxime Pelcat\hspace*{3mm}
Erwan Nogues\hspace*{3mm}
Densen  Puthussery \hspace*{3mm}
Hrishikesh P S\hspace*{3mm}
Jiji C V\hspace*{3mm}
Ashish Sinha \hspace*{3mm}
Xuan Zhao 
}
}
\affil[]{}

\maketitle

\begin{abstract}
This paper reviews the Challenge on Image Demoireing that was part of the New Trends in Image Restoration and Enhancement (NTIRE) workshop, held in conjunction with CVPR 2020. Demoireing is a difﬁcult task of removing moire patterns from an image to reveal an underlying clean image. The challenge was divided into two tracks. Track~1 targeted the \emph{single image} demoireing problem, which seeks to remove moire patterns from a single image.  Track~2 focused on the \emph{burst} demoireing problem, where a set of degraded moire images of the same scene were provided as input, with the goal of producing a single demoired image as output. The methods were ranked in terms of their fidelity, measured using the peak signal-to-noise ratio (PSNR) between the ground truth clean images and the restored images produced by the participants' methods. The tracks had 142 and 99 registered participants, respectively, with a total of 14 and 6 submissions in the final testing stage. The entries span the current state-of-the-art in {\em image and burst image demoireing} problems.
\end{abstract}

\section{Introduction}
\label{report:intro}

{\let\thefootnote\relax\footnotetext{S. Yuan (shanxin.yuan@huawei.com, Huawei Noah's Ark Lab), R. Timofte, A. Leonardis and G. Slabaugh are the NTIRE 2020 challenge organizers, while the other authors participated in the challenge. 
    \\Appendix~\ref{sec:appendix} contains the authors' teams and affiliations.
    \\NTIRE 2020 webpage:\\~\url{https://data.vision.ee.ethz.ch/cvl/ntire20/}}}

Digital photography has matured in the past years with the new advancements in relevant research areas, including image denoising~\cite{DnCNN,gu2019brief,abdelhamed2019ntire}, image demosaicing~\cite{gharbi2016deep}, super-resolution~\cite{romano2016raisr,timofte2017ntire,Timofte_2018_CVPR_Workshops,blau20182018,cai2019ntire,lugmayr2019aim, TGA2020}, deblurring~\cite{nah2019ntire}, dehazing~\cite{ancuti2018ntire}, quality mapping~\cite{Ignatov_2017_ICCV}, automatic white balance~\cite{FFCC}, and high dynamic range compression~\cite{hdrnet}. 
Moire aliasing is a less addressed and fundamental problem. In the case of digital photography, moire aliasing occurs when the camera's color filter array (CFA) interferes with high frequency scene content close to the resolution of CFA grid. The high frequency regions result in undersampling on the sensor color filter array (CFA) and when demosaiced, can create disruptive colorful patterns that degrade the image.  For example, moire aliasing is likely to happen when taking pictures of clothing or long-distance building's tiles.


The AIM19 demoireing challenge~\cite{AIM19demoireMethods} addressed a more specific scenario of photography of digital screens. In this scenario, moire patterns appear when the CFA interferes with the LCD screen's subpixel layout. However, in the photography natural scenes, moire aliasing is also a less addressed problem, where the CFA interferes with high frequency scene content. Our NTIRE2020 demoireing challenge addresses this more general case. 
 
By engaging the academic community with this challenging image enhancement problem, a variety of methods have been proposed, evaluated, and compared using a common dataset.
In this paper, we also propose a new dataset called~\emph{CFAMoire} consisting of 11,000 image pairs (image with moire patterns in the high frequency areas, and clean ground truth).

This challenge is one of the NTIRE 2020 associated challenges on: deblurring~\cite{nah2020ntire}, nonhomogeneous dehazing~\cite{ancuti2020ntire}, perceptual extreme super-resolution~\cite{zhang2020ntire}, video quality mapping~\cite{fuoli2020ntire}, real image denoising~\cite{abdelhamed2020ntire}, real-world super-resolution~\cite{lugmayr2020ntire}, spectral reconstruction from RGB image~\cite{arad2020ntire} and demoireing.

\begin{figure*}[t]
	\scriptsize
	\centering

		\begin{tabular}{ccccccccc}

			\includegraphics[width=0.12\textwidth, height=0.12\textwidth]{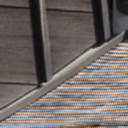}
			\includegraphics[width=0.12\textwidth, height=0.12\textwidth]{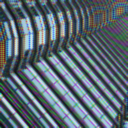}
			\includegraphics[width=0.12\textwidth, height=0.12\textwidth]{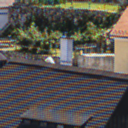}
			\includegraphics[width=0.12\textwidth, height=0.12\textwidth]{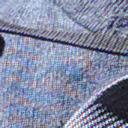}
    		\includegraphics[width=0.12\textwidth, height=0.12\textwidth]{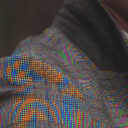}
			\includegraphics[width=0.12\textwidth, height=0.12\textwidth]{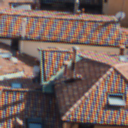}
			\includegraphics[width=0.12\textwidth, height=0.12\textwidth]{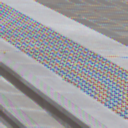}
			\includegraphics[width=0.12\textwidth, height=0.12\textwidth]{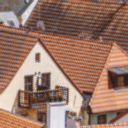}\\
			
			\includegraphics[width=0.12\textwidth, height=0.12\textwidth]{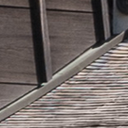}
			\includegraphics[width=0.12\textwidth, height=0.12\textwidth]{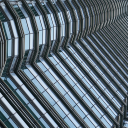}
			\includegraphics[width=0.12\textwidth, height=0.12\textwidth]{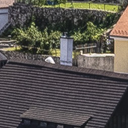}
			\includegraphics[width=0.12\textwidth, height=0.12\textwidth]{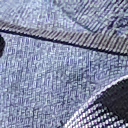}
    		\includegraphics[width=0.12\textwidth, height=0.12\textwidth]{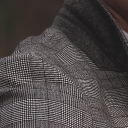}
			\includegraphics[width=0.12\textwidth, height=0.12\textwidth]{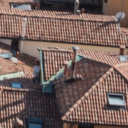}
			\includegraphics[width=0.12\textwidth, height=0.12\textwidth]{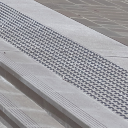}
			\includegraphics[width=0.12\textwidth, height=0.12\textwidth]{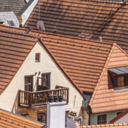}

	\end{tabular}
	
	\caption{Examples from the \emph{CFAMoire} dataset. Top row are moire images, bottom row are ground truth.}
	\label{fig:cfamoireexample}
\end{figure*}

\begin{figure*}[h]
	\scriptsize
	\centering

		\begin{tabular}{ccccccccc}

    		\includegraphics[width=0.12\textwidth, height=0.12\textwidth]{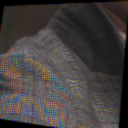}
			\includegraphics[width=0.12\textwidth, height=0.12\textwidth]{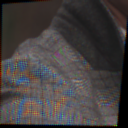}
			\includegraphics[width=0.12\textwidth, height=0.12\textwidth]{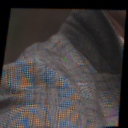}
			\includegraphics[width=0.12\textwidth, height=0.12\textwidth]{000138_3.png}
			\includegraphics[width=0.12\textwidth, height=0.12\textwidth]{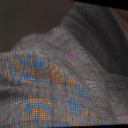}
			\includegraphics[width=0.12\textwidth, height=0.12\textwidth]{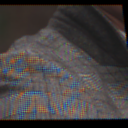}
			\includegraphics[width=0.12\textwidth, height=0.12\textwidth]{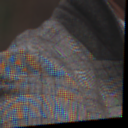}
			\includegraphics[width=0.12\textwidth, height=0.12\textwidth]{000138_gt.png}\\
			
			\includegraphics[width=0.12\textwidth, height=0.12\textwidth]{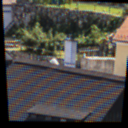}
			\includegraphics[width=0.12\textwidth, height=0.12\textwidth]{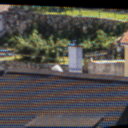}
			\includegraphics[width=0.12\textwidth, height=0.12\textwidth]{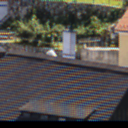}
			\includegraphics[width=0.12\textwidth, height=0.12\textwidth]{000044_3.png}
			\includegraphics[width=0.12\textwidth, height=0.12\textwidth]{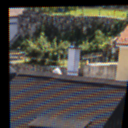}
			\includegraphics[width=0.12\textwidth, height=0.12\textwidth]{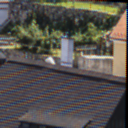}
			\includegraphics[width=0.12\textwidth, height=0.12\textwidth]{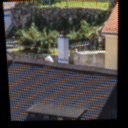}
			\includegraphics[width=0.12\textwidth, height=0.12\textwidth]{000044_gt.png}\\
	
	\end{tabular}
	
	\caption{Two examples for the burst image track. Random homographies are applied to the clean image before the remosaicing and demosaicing step. The last column is the ground truth. With slight viewpoint changes, the moire patterns change dramatically. A burst of images can provide extra information to the reference image.}
	\label{fig:track2examples}
\end{figure*}

\section{The Challenge}
\label{report:challenge}

The demoireing challenge was hosted jointly with the \textit{New Trends in Image Restoration and Enhancement (NTIRE)} workshop held in conjunction with the IEEE Computer Society Conference on Computer Vision and Pattern Recognition (CVPR) 2020.

The task of image demoireing is to design an algorithm to remove moire patterns from an input image.  To help build algorithms, particularly those based on machine learning, a novel dataset was produced for the challenge.  

\subsection{CFAMoire Dataset}
As an essential step towards reducing moire effects, we proposed a novel dataset, called \emph{CFAMoire}. 
It consists of 10,000 training, 500 validation, and 500 test images with 128 $\times$ 128 resolution.
The images are high quality in terms of the reference frames and different moire patterns. It also covers balanced content, including clothing and buildings, where the high frequency repetitive patterns interfere with the camera’s color filter array.

The clean images are sampled or cropped from existing datasets, including DeepJoint \cite{gharbi2016deep}, HDR+ \cite{hasinoff2016burst}, and DIV8K \cite{gu2019div8k}. The corresponding moire images are generated through a remosaicing and demosaicing process, where moire artifacts will appear in the high frequency area. 
To collect the clean images, we followed a two-step strategy through first automatically selecting images by quantification of moire artifacts in the Fourier domain, followed by  manual selection. 
The automatic selection step is conducted by measuring the frequency change between the clean image and the demosaiced image \cite{gharbi2016deep}. 

Assuming the clean image and the demosaiced image to be $I_{c}$ and $I_{m}$. Both images are first converted to the Lab space and then a 2D Fourier transform is applied to each channel to get $F(I_{c})$ and $F(I_{m})$. The frequency ratio is calculated by 

\begin{equation}
  \rho(w)=\begin{cases}
               log(\frac{|F(I_{m})|^{2} + \eta}{|F(I_{c})|^{2} + \eta}) \quad\quad if \quad|w|<\gamma\\
               1 \qquad\qquad\qquad\qquad otherwise
            \end{cases}
\end{equation}
where $w$ is the spatial frequency and $\gamma$ is a threshold.  Following the setting of \cite{gharbi2016deep}, we only compare the ratio in frequencies lower than $\gamma$, which is set to $0.95\pi$, to mitigate high-frequency noise. The ratio map is then smoothed with Gaussian blur. If the maximum ratio value across all channels and frequencies exceeds a threshold of 2, then the image is selected.

\begin{table*}[t]
\normalsize 
 \resizebox{\linewidth}{!}{
  \begin{tabular}{llllllllll}
  \toprule 
  \bf Team & \bf PSNR  & \bf SSIM & \bf Time (train, test) & \bf GPU & \bf Platform & \bf Loss &\bf Ensemble & \bf GFlops & \bf \#Para \\
  \midrule 

HaiYun
& 42.14
& 0.99
& 3 days, 0.45
& GTX Titan XP
& Pytorch 1.0.0
& L1 (RGB) + L1 (DCT)
& Self-ensemble
& 45.17
& 144.33M
\\

OIerM
& 41.95  
& 0.99 
& 67 hours, 0.1
& RTX 2080Ti 
& PyTorch 
& L1
& Refinement x16
& - 
& 56.2M \\

Alpha
& 41.84
& 0.99
& 5 days, 0.34
& RTX 2080Ti
& PyTorch
& L1 charbonnier
& Self-ensemble x8
& 267
& 23M
\\

Reboot
& 41.11  
& 0.99 
& 10 days, 0.87 
& RTX 2080Ti
& PyTorch 1.4
& L1 (RGB) + L1 (UV)
& Self-ensemble x4
& -
& 164M\\

LIPT
& 41.04
& 0.99 
& 7 days, 1.5 
& Tesla V100
& TF 2.0
& L1 (RGB) + L1 (${\nabla}RGB$) + L1 (DCT)
& Self-ensemble x8
& - 
& 5.1M \\

iipl-hdu 
& 40.68  
& 0.99 
& 3 days, 0.1 
& 2080Ti 
& Tensorflow 1.14
& L1 + ASL
& -
& - 
& 24M \\

Mac AI
& 40.60
& 0.99 
& 3 days, 0.1635 
& GTX 1080Ti 
& PyTorch 1.1
& L1 charbonnier + SSIM
& Self-ensemble x8
& - 
& 13.6M \\

MoePhoto
& 39.05 
& 0.99 
& 24 hours, 0.14
& RTX2080Ti (11GB) x4  
& PyTorch 1.4
&  L1 charbonnier + L1 Wavelet 
& None
& 46.38 
& 6.48M \\

DGU-CILab
& 38.28
& 0.98
& 3 days, 0.187
& RTX 2080Ti
& PyTorch 1.3
&  L2
& Self-ensemble x8
& -
& 14M
\\

Image Lab
& 36.73  
& 0.98 
& -,0.01944 
& GPU with 8GB memory
& TF 1.10 
& L2 + SSIM + Sobel
& - 
& - 
& 1.6M \\

VAADER
& 36.63 
& 0.98 
& 2days, 0.035
& 2080Ti train, GTX 1080Ti (test)
& PyTorch 1.4
& L1
& Self-ensemble x8
& - 
& 13M \\

CET\_CVLab
& 32.07 
& 0.95 
& 5 days,0.192
& Nvidia Quadro K6000
& TF 2.1
& L1 + color loss
& No
& - 
& - \\

sinashish
& 30.69 
& 0.92 
& 6 hours, 0.045
& Tesla K80
& PyTorch 1.14
& L2
& No
& - 
& - \\

NTIREXZ
& 29.52  
& 0.90
& 18 hours, 0.115
& Nvidia 1060
& - 
& L1
& 
& - 
& - \\

no processing
& 25.45 
& 0.77
& -
& -
& -
& - 
& 
& - 
& - \\

  \bottomrule
  \end{tabular}
 }
  
    \caption{Final results for the \textbf{Single image} demoire challenge track (Track 1).}
  \label{tab:track1} 
	\vspace{-10pt}
\end{table*}

\begin{table*}[h]
\normalsize 
  \resizebox{\linewidth}{!}{
  \begin{tabular}{llllllllll}
  \toprule 
  \bf Team & \bf PSNR  & \bf SSIM & \bf Time (train, test) & \bf GPU & \bf Platform & \bf Loss  &\bf Ensemble &\bf GFlops &\bf \#Para\\ 
  \midrule

OIerM
& 41.95  
& 0.99 
& 67 hours, 0.1
& RTX 2080Ti 
& PyTorch 
& \bf Loss 
& Refinement x16
& 
& 56.2M \\

Alpha
& 41.88
& 0.99
& 3 days, 0.34
& RTX 2080Ti
& Pytorch
& L1 charbonnier
& Self-ensemble x8
\\

Alpha$^*$
& 45.32
& 1.00
& 5 days, 0.34
& RTX 2080Ti
& PyTorch
& L1 charbonnier
& Self-ensemble x8
\\

Mac AI
& 40.64 
& 0.99  
& 3 days, 0.2753 
& GTX 1080Ti 
& PyTorch 1.1 
& L1 charbonnier + SSIM 
& Self-ensemble x8
& - 
& 13.8M \\

Reboot
& 40.33  
& 0.99 
& 10 days, 0.22
& RTX 2080Ti
& PyTorch 1.4
& L1 (RGB) + L1 (UV)
& None
& - 
& 21M\\

MoePhoto
& 39.05 
& 0.99 
& 24 hours, 0.14
& RTX 2080T x4  
& PyTorch 1.4
&  L1 charbonnier + L1 Wavelet 
& None
& 46.38
& 6.48M \\

DGU-CILab
& 38.50
& 0.99
& 4 days, 0.1636
& RTX 2080Ti
& PyTorch 1.3
& -
& Self-ensemble x8
& -
& 14M
\\

no processing
& 25.45 
& 0.77
& -
& -
& - 
& -
& - 
& - 
& - \\

  \bottomrule
  \end{tabular}}
  
    \caption{Final results for the \textbf{Burst}  demoire challenge track (Track 2). $^*$ submission after the deadline.}
  \label{tab:track2} 
  	\vspace{-10pt}

\end{table*}

This technique can produce a large number of candidate image pairs for the dataset. However, it cannot detect clean images that are already corrupted with moire artifacts. Moreover, since it works in the Fourier domain and requires some manually set parameters, it may select suboptimal image pairs.
Therefore, the organizers spent one week on manual selection.  As a result, only 10\% of the pairs from the automatic step are selected. Figure \ref{fig:cfamoireexample} shows some examples.

\subsection{Tracks and Evaluation}
The challenge had two tracks: \textbf{Single image} track and \textbf{Burst} track.

\textbf{Single image:} In this track, participants developed demoire methods to achieve a moire-free image with the best fidelity compared to the ground truth. Figure \ref{fig:cfamoireexample} shows some image pairs.

\textbf{Burst:} Similar to the Single image track, this track works on a burst of images. To generate the input data, apart from the reference moire image, 6 supporting images are generated by first applying random homographies to the clean image and then by going through the mosaicing and demosaicing step, see Figure \ref{fig:track2examples} for examples. 

For both tracks, we used the standard Peak Signal To Noise Ratio (PSNR) measure to rank methods.  Additionally, the Structural Similarity (SSIM) index was computed. Implementations of PSNR and SSIM are found in most image processing toolboxes. For each method we report the average results over all the processed images.

\subsection{Competition}

\textbf{Platform:} The CodaLab platform was used for this competition. To access the data and submit their demoired image results to the CodaLab evaluation server each participant had to register.

\textbf{Challenge phases:} 
(1) Development (training) phase: the participants received both moire and moire-free training images of the dataset.
(2) Validation phase: the participants had the opportunity to test their solutions on the moire validation images and to receive immediate feedback by uploading their results to the server. A validation leaderboard was available. 
(3) Final evaluation phase: after the participants received the moire test images and clean validation images, they submitted their source code / executable, demoired images, and a factsheet describing their method before the challenge deadline. One week later, the final results were made available to the participants.

\section{Results}
\label{report:results}

For the single image track, there were 142 registered teams, and 14 teams entered the the final phase by submitting results. Table~\ref{tab:track1} reports the final test results, rankings of the challenge, self-reported runtimes and major details from the factsheets. The leading entry was from the HaiYun team, scoring a PSNR of $42.14$ dB.  Second and third place entries were by teams OIerM and Alpha respectively.

For the burst track, there were 99 registered teams, and 6 teams entered the final phase by submitting results. Table~\ref{tab:track2} reports the results and details from the factsheets. The leading entry was from the OIerM team, scoring a PSNR of $41.95$.  Second and third place entries were by teams Alpha and Mac AI respectively. Please note that the best PSNR in the burst track is \emph{lower} than that of single frame track, which is counter-intuitive.  One would expect the additional frames to provide more data useful to produce a better quality result. However, after the challenge's deadline, Alpha team submitted new results that significantly increased the PSNR to 45.32 (larger than the best result on single frame track) by making use of alignment across the burst.

Looking across entries, some interesting trends were noted.  In particular,
\begin{itemize}
\item \textbf{Ensembles:} Most solutions used self-ensemble~\cite{SevenWays} x8. Small improvements are reported by the solutions. Besides self-ensemble, a new fusion method is also proposed by OIerM team. 
OIerM team performed 16 invertible transforms including rotating, flipping, transposing and their combinations on a single image. Then they perform corresponding inverse transforms on the outputs and concatenated them together to get augmented results. Then they train a smaller network to fuse the augmented results.
This strategy brings +0.35dB PNSR over the single method on validation set.

\item \textbf{Multi-scale strategy:} Most solutions adopted a multi-scale strategy as a mechanism to handle moire patterns of different frequencies.

\item \textbf{New loss functions}: Although many solutions choose the traditional L1 loss function. A few new loss functions are used, including coral loss by HaiYun team, L1 Wavelet loss by MoePhoto team, color loss by CET\_CVLab. 
\end{itemize}

The next section describes briefly the method of each team, while in the Appendix A the team members and their affiliations are provided.

\section{Challenge Methods}

\subsection{HaiYun team}

\begin{figure}[h]
	\centering
	\includegraphics[width=0.45\textwidth]{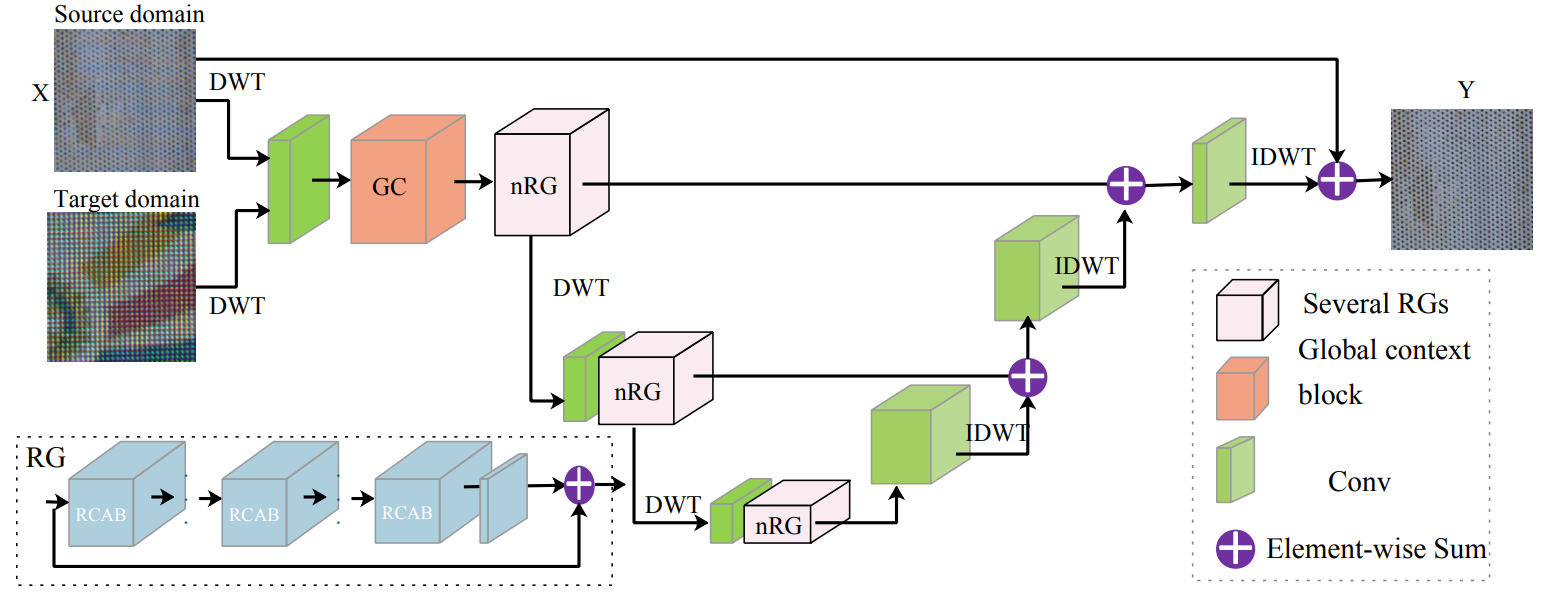}
	\vspace{-10pt}
	\caption{The overall network architecture of the proposed AWUDN.}
	\label{fig:network_architecture}
	\vspace{-10pt}
\end{figure}

\begin{figure}[t]
	\centering
	\includegraphics[angle=90, width=0.25\textwidth, height=0.2\textwidth]{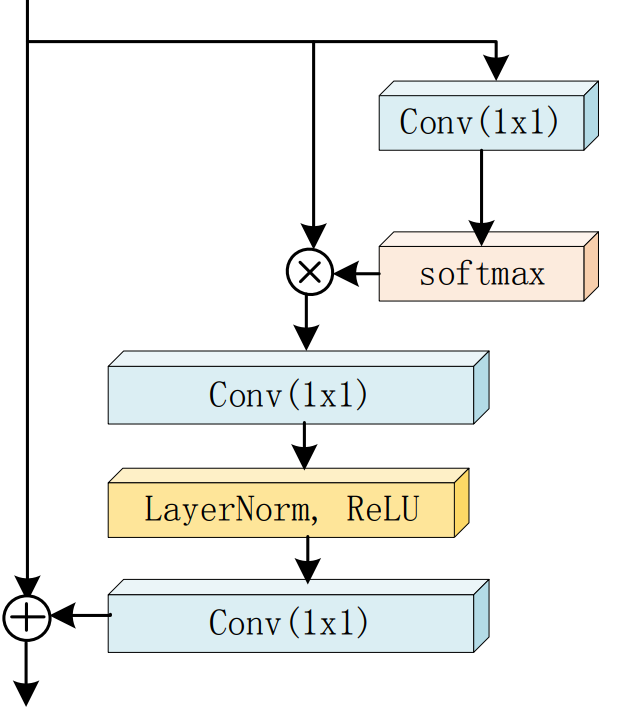}
	\caption{Global context block}
	\label{fig:non-local}
	\vspace{-15pt}
\end{figure}

\textbf{HaiYun} team propose a deep wavelet network with domain adaptation for single image demoireing, dubbed AWUDN \cite{Luo_2020_CVPR_Workshops}, as shown in Figure \ref{fig:network_architecture}. The whole network is an U-Net structure, where the downsampling and upsampling of feature maps are replaced with discrete wavelet transform (DWT) and inverse discrete wavelet transform (IDWT) for reducing computation complexity and information loss. Therefore, the feature mapping is performed in wavelet domain, where the basic block adopts the residual-in-residual structure \cite{zhang2018image} for extracting more residual information effectively. Considering that the dataset provided by the competition has strong self-similarity, i.e., similar texture structure, the global context block \cite{gcnet} as shown in Figure \ref{fig:non-local} is introduced in the front of network structure. It can help establish the relationship between two distant pixels to better use the internal information of the image for restoring texture details. Moreover, there may exist slight domain difference between the source domain training data and the target domain testing data. It means that the distribution of moire images in the training set and the moire images in the test set is inconsistent, which can constrain the performance improvement of the model pretrained on the training dataset. CORAL loss \cite{coral} gives us some inspiration, which defines the measurement difference of the second-order statistics between features in the source domain and the target domain. Therefore, the pretrained model WUDN is fine-tuned using coral loss for reducing the domain shift of training moire dataset and testing moire dataset in the testing phase. Self-ensemble strategy is adopted to further improve performance. The proposed solution can obtain significant quantitative and qualitative results.

\subsection{OIerM team}
\textbf{OIerM} team propose an Attentive Fractal Network (AFN)~\cite{OIerM_2020_Demoire} to effectively solve the demoir{\'e} problem. First, they construct an Attentive Fractal Block via progressive feature fusion and channel-wise attention guidance, then they stack our AFB in a fractal way inspired by FBL-SS~\cite{yang2020derain}. Finally, to further boost the performance, they adopt a two-stage augmented refinement strategy.

\begin{figure}[t]
  \centering
  \includegraphics[width=0.45\textwidth]{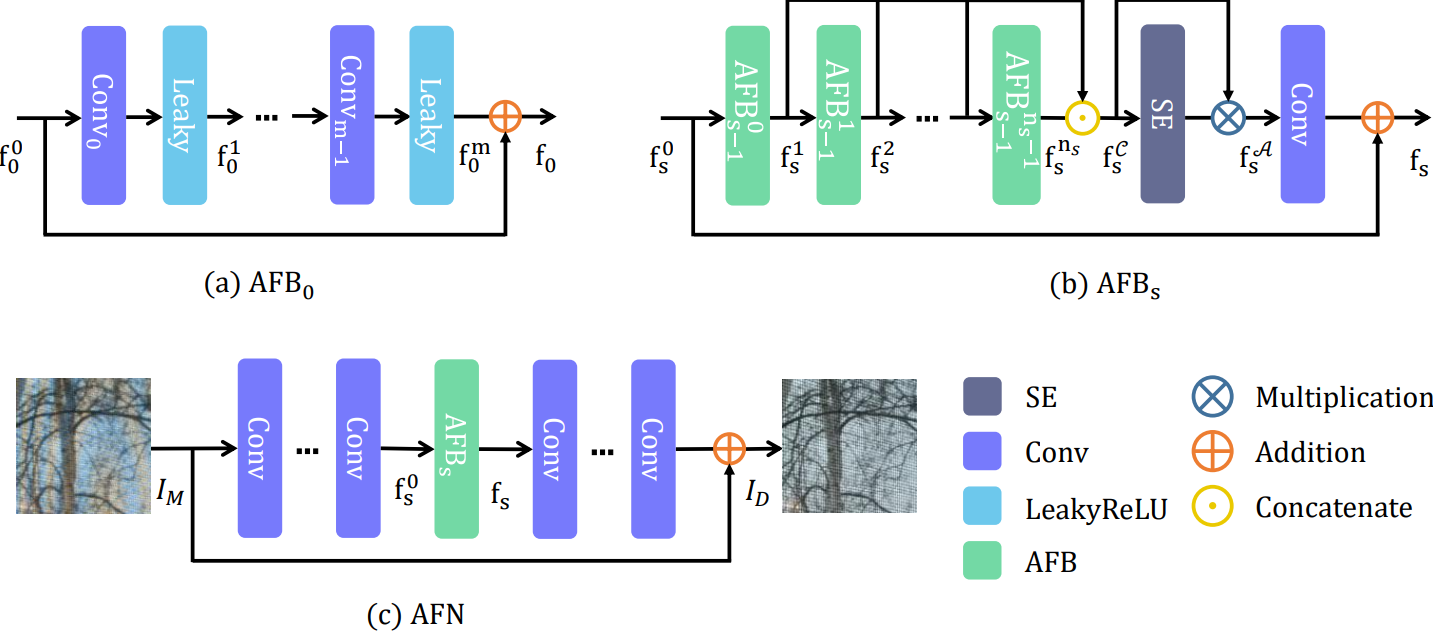}
  \caption{Architecture of AFN. (a) shows the structure of the basic block $\text{AFB}_0$. (b) shows the hierarchical architecture of $\text{AFB}_s$, where $s$ is the recursion level of the block. (c) illustrates the structure of AFN, which consists of an encoding block, an $\text{AFB}_s$, and an decoding block.}
  \label{fig:framework}
\end{figure}

The proposed AFN adopts a fractal network architecture. With the help of shortcuts and residuals of different levels, the whole network gains the ability to utilize both local and global features for moir{\'e} pattern removal. The framework, as shown in Fig.~\ref{fig:framework} (c), consists of three main parts, which are encoding layers $G_\text{E}$, an Attentive Fractal Block (AFB) $G_\text{AFB}$, and decoding layers $G_\text{D}$.

The encoding layers first transform the input image $I_M$ into a multi-channel feature $f_s^{0}$, which is the input of the following AFB of level s.
\begin{equation}
  f_s^0 = G_\text{E}(I_{M}).
\end{equation}
The AFB performs major refinement for the encoded features to obtain $f_s$.
\begin{equation}
  f_{s}=G_{\text{AFB}_s}(f_s^0).
\end{equation}
The feature $f_s$ is then given to the decoding layers to reconstruct a three-channel clean image $I_D$. A global residual connection is added to stabilize the network.
\begin{equation}
  I_D=G_\text{D}(f_s)+I_M.
\end{equation}
They also adopt a two-stage augmented refinement strategy to push the ability of AFN further. Specifically, the second stage uses a similar but shallower AFN network with lower levels of AFB to refine the output of the first stage.

\noindent \textbf{Attentive Fractal Block.} The proposed AFB is built in a fractal way, and it has a self-similar structure. Each high-level AFB can be constructed with AFBs of lower-level recursively until the level reaches zero. Specifically, as shown in Fig.~\ref{fig:framework}(b), an AFB of level $s$ consists of $n_s$ AFBs of level $s-1$ as well as a Fusion Unit. In Fig.~\ref{fig:framework}(a), a level $0$ AFB is illustrated, and it has a residual shortcut and $m$ convolution layers followed by LeakyReLU layers.
For level $s$, the encoded features $f_s^0$ are fed into each of the ${n_s}$ AFBs of level $s-1$ sequentially to obtain ${n_s}$ refined features, which are $f_s^1, \cdots, f_s^{n_s}$ respectively:
\begin{equation}
  f_{s}^{i}=G_{\text{AFB}_{s-1}^{i-1}}(f_{s}^{i-1}), (1 \leq i \leq n_s).
\end{equation}
They then adopt a progressive feature fusion strategy, which is to progressively feed-forward the intermediate features of early stages to the end of the current block and fusion them there. By concatenating the features of the same level, their rich information from different stages help the network learn more thoroughly:
\begin{equation}
  f_{s}^{\mathcal{C}}=[f_s^1,f_s^2, \cdots ,f_s^{n_s}].
\end{equation}
Too many channels might confuse the network with abundant information, so they choose to adopt channel-wise attention with the help of a Squeeze-and-Excitation Layer~\cite{hu2018squeeze}. By multiplying the assigned learnable weights, the output feature maps are re-weighted explicitly according to their properties.
\begin{equation}
  f_{s}^{\mathcal{A}}=G_{\text{SE}_s}(f_{s}^{\mathcal{C}}) \cdot f_{s}^{\mathcal{C}}.
\end{equation}
Finally, they send the features to a Fusion Unit (FU) to narrow down their channels. A local residual connection is also adopted to stabilize the network.
\begin{equation}
  f_{s}=G_{\text{FU}_s}(f_{s}^{\mathcal{A}})+f_{s}^0.
\end{equation}
For level $0$ AFB, it sends the input $f_0^0$ sequentially to $m$ convolution layers followed by LeakyReLU layers.
\begin{equation}
  f_0^{i}=F_\text{Leaky}(G_{\text{Conv}_i}(f^{i-1}_0)), (1 \leq i \leq m).
\end{equation}
A local shortcut is also adopted to relieve the burden of the network.
\begin{equation}
  f_{0}=f_0^{m}+f_0^{0}.
\end{equation}
As illustrated above, a high-level AFB is made up of several lower-level AFBs. So actually the input of a level $s-1$ AFB $f_{s-1}^0$ is also the input $f_{s}^i$ of an intermediate layer in a level $s$ AFB. And the output of a level $s-1$ AFB $f_{s-1}$ is also the output $f_{s}^{i+1}$ of an intermediate layer in a level $s$ AFB.

\subsection{Alpha team}

\begin{figure}[t]
	\vspace{-5pt}
	\centering
    \includegraphics[width=0.450\textwidth]{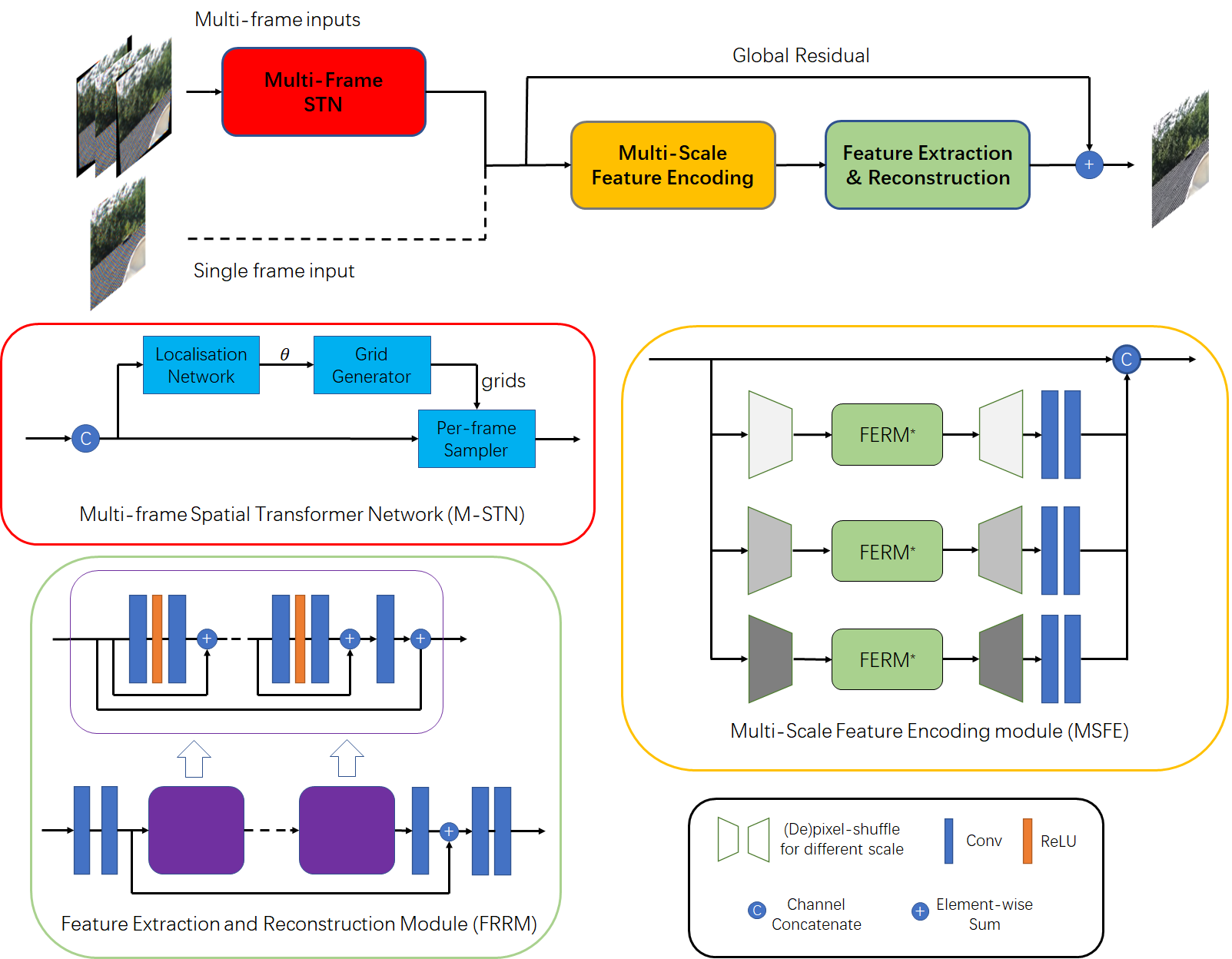}
	\caption{Network architecture of multi-frame and multi-scale for image demoireing (MMDM).}
	\label{fig:alphaframework}
		\vspace{-10pt}
\end{figure}

\begin{figure}[h]
	\centering
    \includegraphics[width=0.35\textwidth]{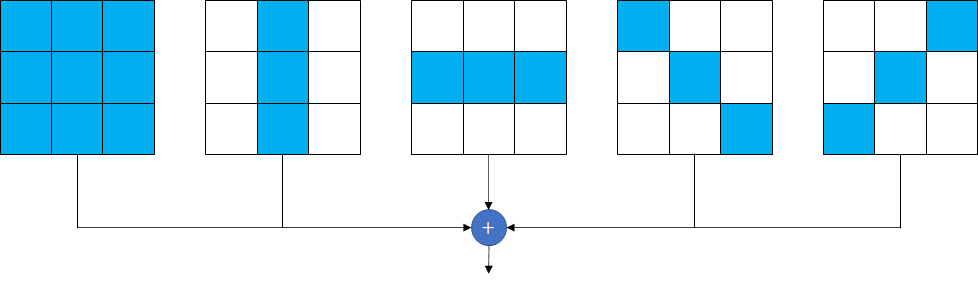}
	\caption{Enhanced asymmetric convolution block (EACB).}
	\label{fig:EAC}
	\vspace{-10pt}
\end{figure}

\textbf{Alpha team} propose the \textit{MMDM: Multi-frame and Multi-scale for Image Demoireing} \cite{Liu2020MMDM}, see Figure \ref{fig:alphaframework}. They propose a feature extraction and reconstruction module (FERM) based on RCAN \cite{zhang2018image} by removing all channel attention modules and upsampling layers, and adding a global residual connection. FERM has less inference time and more stable output. 
In order to process the various frequency components in the moire patterns, they propose a multi-scale feature encoding module (MSFE) that processes images at different scales. The MSFE has 3 simple versions of FERM with up and down sampling layers for different scales.
In addition, they also propose an enhanced asymmetric convolution block (EACB), which can be applied to any image restoration network, see Figure \ref{fig:EAC}. Based on ACB \cite{ding2019acnet}, they add two additional diagonal convolutions to further strengthen the kernel skeleton, and remove the batch normalization layers and the bias parameters for better performance. 
They used the proposed FERM and EACB to achieve the second place in the track 2: sRGB of the real image denoising challenge \cite{abdelhamed2020ntire}.

For the burst task, they propose a multi-frame spatial transformer network (M-STN). In order to connect multiple input frames, they concatenate $n$ input frames together over the channel dimension ($n$ is the number of input frames, including 1 standard frame and $n-1$ non-aligned frames). The localisation network processes the input to get $8\times (n-1)$ parameters (they use the perspective transform which can simulate the burst input, and only transform the non-aligned frames), and constructs $n-1$ perspective transformation matrices. The grid generator and sampler respectively transform the input frames according to the matrices to obtain $n$ aligned frames. The aligned frames are concatenated together over the channel dimension, then input to the main network. With more aligned high-frequency information, the performance of the network has been greatly improved. Compared to STN \cite{jaderberg2015spatial}, the M-STN can process multi-frame inputs at the same time and make information fusion between them.

For both tracks, the team only used the simplified version of EACB (without diagonal convolutions) because EACB had a long training time.
The use of EACB in this model, although it will consume more time during the training phase, the asymmetric convolutions can be fused, and there is no additional time consumption during testing phase.

\textbf{Ensembles and fusion strategies:} 
Both tracks use self-ensemble X8 (flip and rotate). After training, the weight of the asymmetric convolution is added to the corresponding position to achieve the fusion effect. This is not like a general fusion strategy, but more like self-fusion.

\subsection{Reboot team}

\begin{figure}[h]
	\begin{center}
		\includegraphics[width=0.45\textwidth]{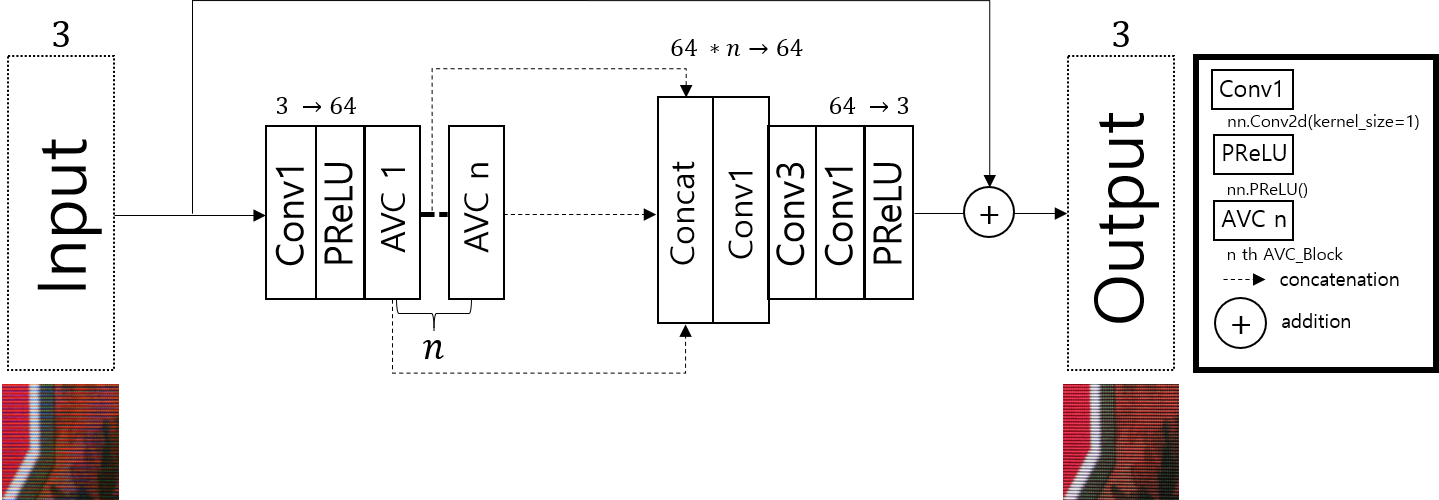}
	\end{center}
	\vspace{-10pt}
	\caption{Overall structure of C3Net.}
	\label{fig:msfan}
	\vspace{-10pt}
\end{figure}    

\begin{figure}[h]
	\begin{center}
		\includegraphics[width=0.45\textwidth]{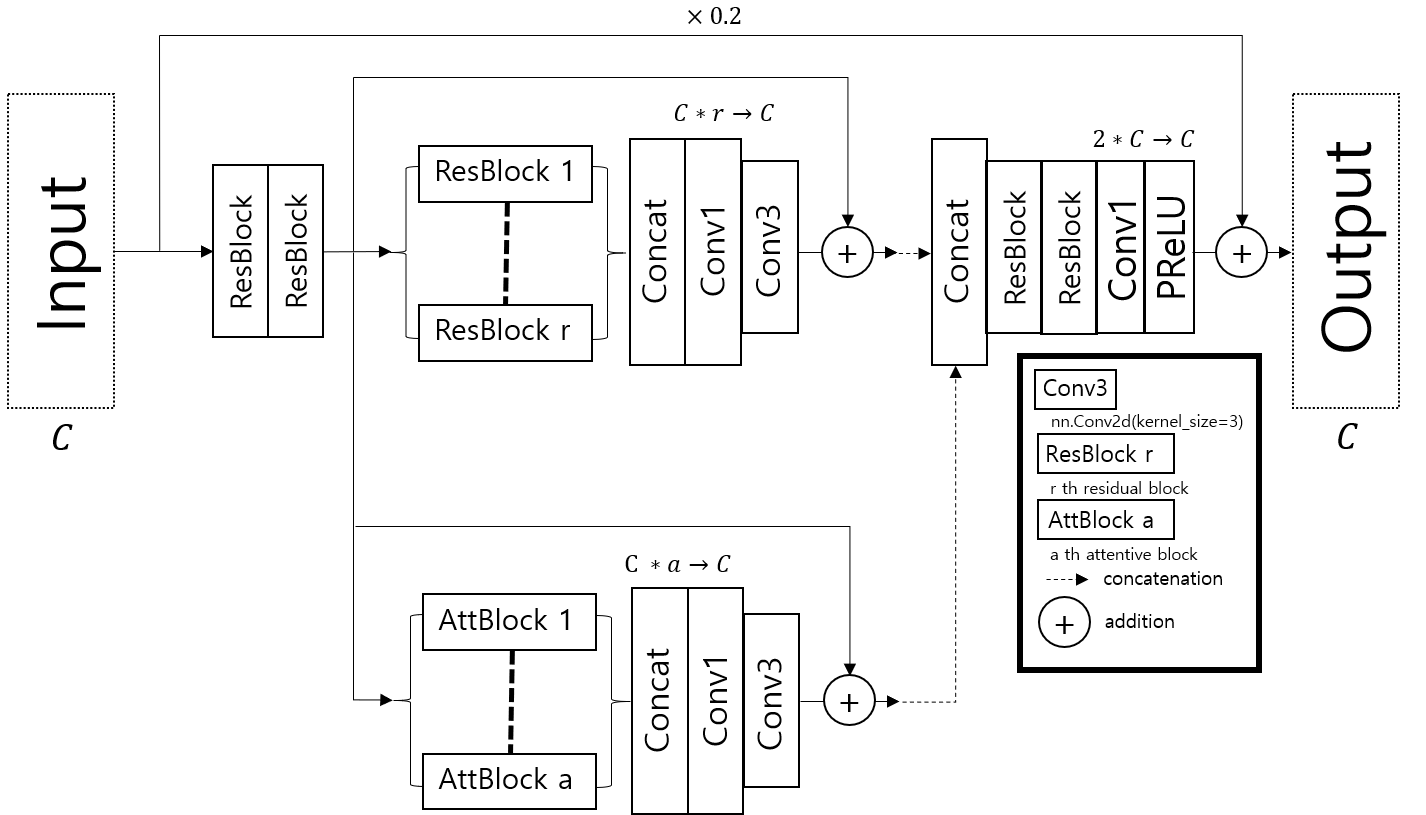}
	\end{center}
	\vspace{-10pt}
	\caption{The structure of Attention-Via-Concat Block (AVCBlock).}
	\label{fig:avcblock}
	\vspace{-15pt}
\end{figure}

The \textbf{Reboot} team proposed \textit{C3Net: Demoireing Network Attentive in Channel, Color and Concatenation} \cite{kim2020c3net}.
The method is inspired by Residual Non-local Attention Network (RNAN) \cite{zhang2019residual} and Deep Iterative Down-Up CNN for Image Denoising (DIDN) \cite{yu2019deep}. 
The entire network consists of $n$ Attention Via Concatenation Blocks (AVCBlocks) in global feature fusion used in Residual Dense Network for Image Super-Resolution (RDN)\cite{zhang2018residual}, see Figures \ref{fig:msfan}. 
An AVCBlock, see Figure \ref{fig:avcblock}, consists of two branches: the trunk branch and the mask branch. In the trunk branch, there are $r$ residual blocks (ResBlocks) in  parallel to retain the original values of input in diverse ways. In the mask branch, there are $a$ attentive blocks (AttBlock) for guiding the values from trunk branch to demoire. Outputs from the two branches are concatenated and the number of channels of concatenated features is halved for entering next block.
The method uses ResBlocks, each ResBlock similar to the one used in \cite{yu2019local} which consists of one convolutional layer, one PReLU layer, and another convolutional layer. Channel attention \cite{zhang2018image} is added with ReLU to cope with demoireing problems related to colors. 
An AttBlock is a U-net with Resblocks, convolutional layers with stride 2 for downscaling, and pixel shuffle with kernel size 2 for upscaling features. The scaling layers and usage of U-net as a block benchmarked DIDN.

For burst processing, global maxpooling \cite{aittala2018burst} is added following each AVCBlock. The algorithm gives feature maps which have maximum among 7 images and replicate and concatenate 7 times to match the dimension. 
The proposed C3Net concatenates the output of AVCBlock and the output of AVCBlock and global maxpooling layer and the number of channels of concatenated features halves for entering next block.

\subsection{LIPT}

\begin{figure}[h]
	\centering
	\includegraphics[width=1.0\linewidth]{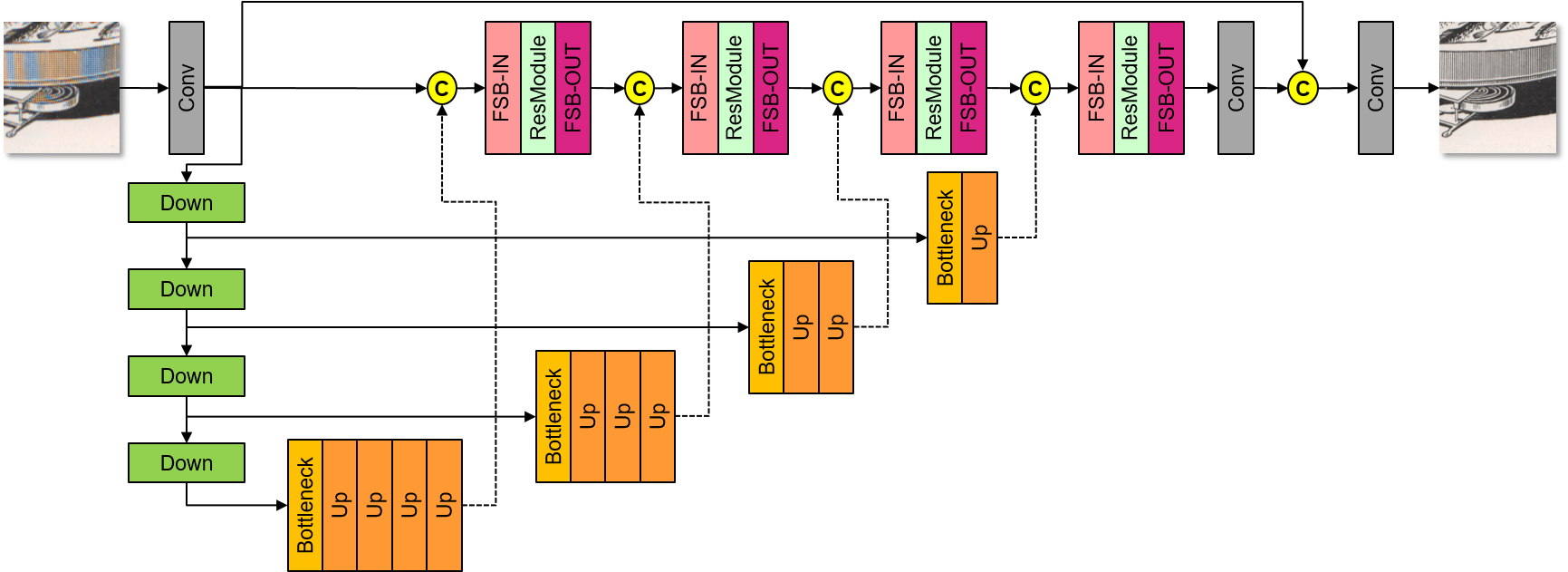}
	\caption{Overall architecture of the multi-scale deep residual network with adaptive pointwise convolution model.}
	\label{fig:modelarchitecture}
\end{figure}

\begin{figure}[h]
	\centering
		\vspace{-10pt}
	\begin{subfigure}{.15\textwidth}
		\includegraphics[width=\linewidth]{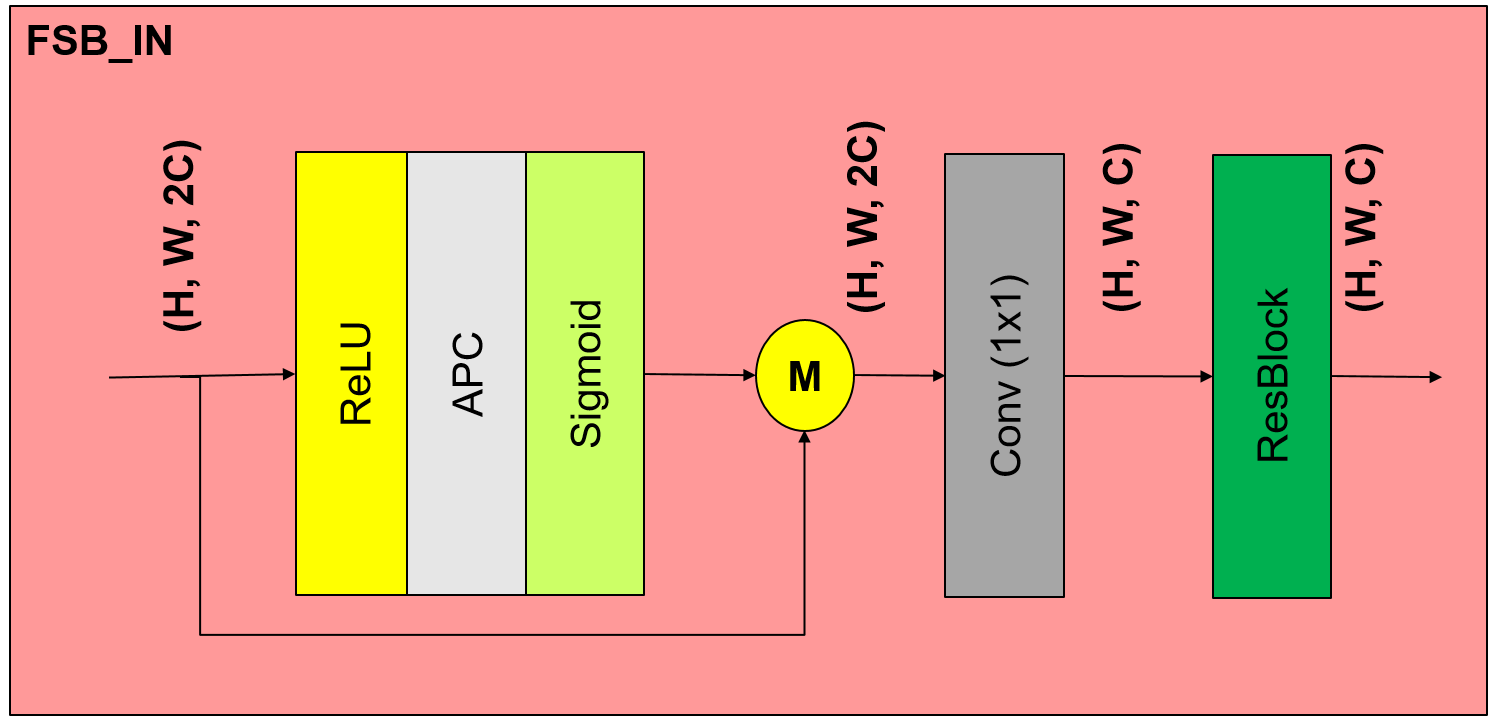}
		\caption{\textit{FSB\_IN}}
		\label{fig:Fig_FSB_IN}
	\end{subfigure}
	\begin{subfigure}{.15\textwidth}
		\includegraphics[width=\linewidth]{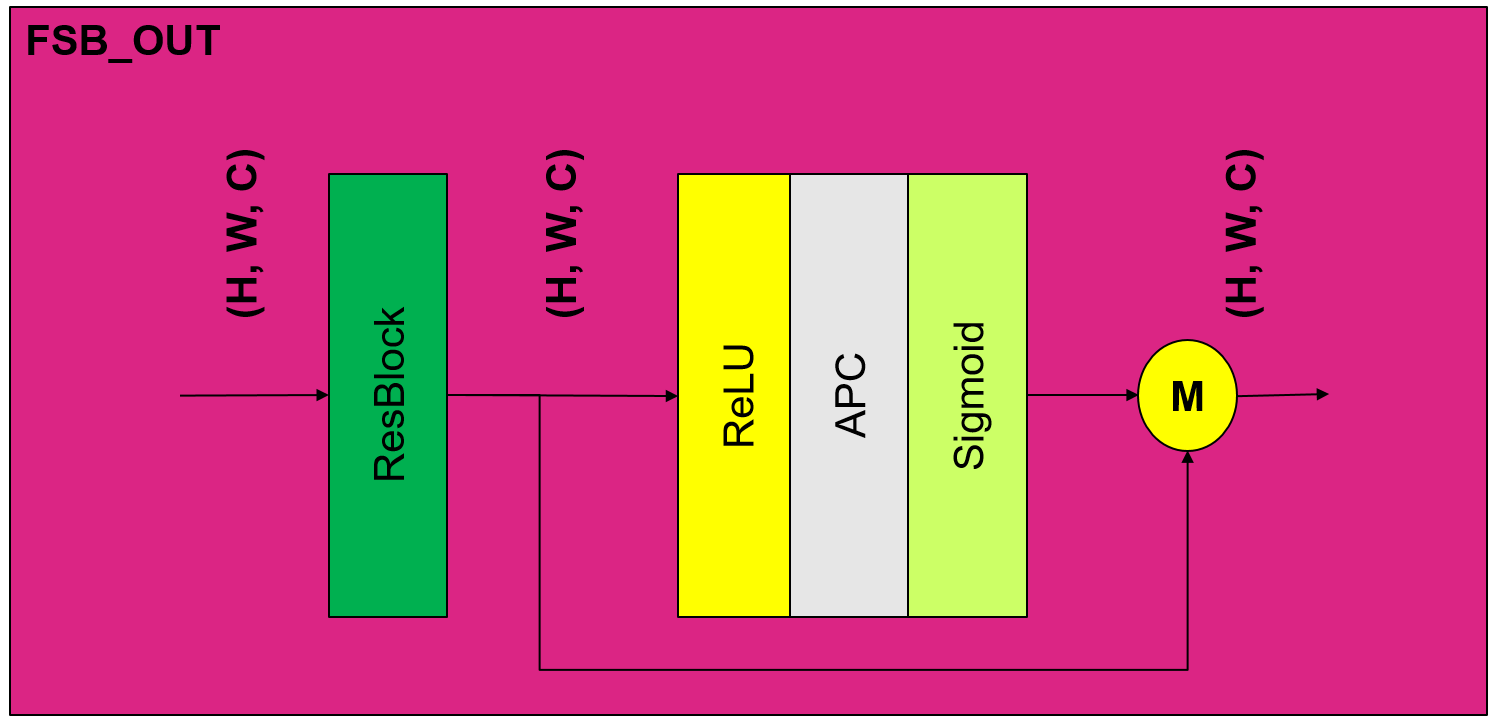}
		\caption{\textit{FSB\_OUT}}
		\label{fig:Fig_FSB_OUT}
	\end{subfigure}
	\begin{subfigure}{.15\textwidth}
		\includegraphics[width=\linewidth]{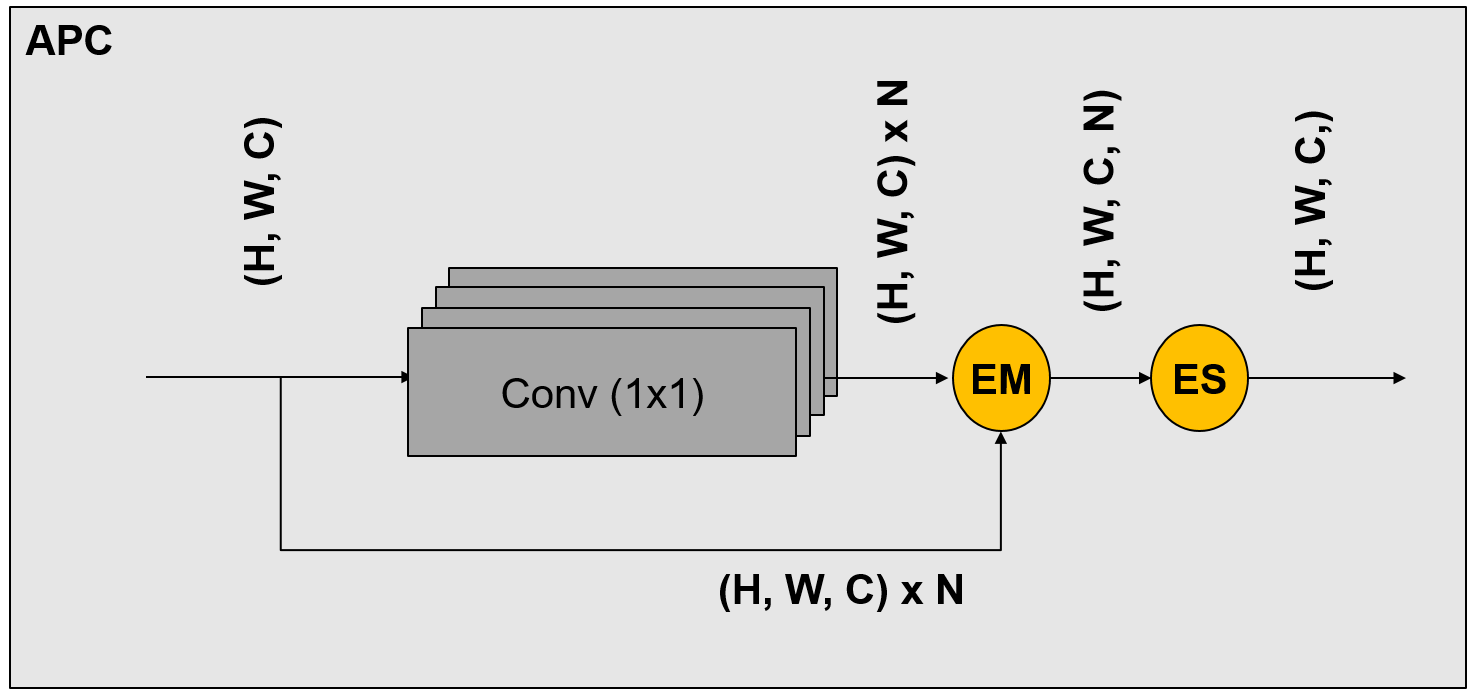}
		\caption{\textit{APC}}
		\label{fig:Fig_APC}
	\end{subfigure}
	\caption{Proposed modules in the model.}
	\label{fig:proposed_modules_LIPT}
	\vspace{-10pt}
\end{figure}

The \textbf{LIPT} team proposed \textit{Multi-scale deep residual network with adaptive pointwise convolution}.
In order to extract more useful features, a novel adaptive pointwise convolution (APC) is proposed and applied, which is a modified version of a pointwise convolution based on 1$\times$1 convolution filters \cite{tian2019decoders}. The APC uses spatially variant learnable kernel weights for each pixel feature in order to perform adaptive linear projection.
As shown in Figure \ref{fig:modelarchitecture}, the overall architecture consists of two branches, namely, main and multi-scale branches. 
Features are extracted from the input image by applying a 3$\times$3 convolution.

Each multi-scale branch performs down-scaling, feature extraction, and up-scaling processes by using \textit{Down}, \textit{Bottleneck}, and \textit{Up} modules, respectively. It can handle multi-scale features efficiently. The \textit{Down} module is a 3$\times$3 convolution with a stride of 2. The \textit{Up} module expands the feature map size by applying a 3$\times$3 convolution and a sub-pixel convolution \cite{shi2016real}. The \textit{Bottleneck} module consists of a pointwise convolution with ReLU as the activation function, the \textit{Up}, and the \textit{Down} modules.

The extracted multi-scale information is merged into the main branch, which consists of several residual blocks (\textit{ResBlocks}).
Four-level multi-scale branches are used, thus, the \textit{ResBlocks} are grouped into 4 \textit{ResModules}.
In order to extract important features from the main and multi-scale branches, feature selection blocks (\textit{FSBs}) are used at the beginning and end of the each \textit{ResModule}.
The proposed \textit{FSB\_IN} and \textit{FSB\_OUT} are shown in Fig. \ref{fig:proposed_modules_LIPT}.
These include the \textit{ResBlock} and the feature selection unit.
The feature selection unit is modified from the selection unit \cite{choi2017deep}.
The proposed \textit{APC} is used instead of a 1$\times$1 convolution in the selection unit.
It can help to extract useful features from the main and multi-scale branches along channels as a channel attention.

\textbf{Adaptive pointwise convolution:} Adaptive convolution has adaptive kernel weights that are learnable for each pixel feature \cite{niklaus2017video2, su2019pixel}. 
A new adaptive pointwise convolution (\textit{APC}) is proposed, as shown in Figure \ref{fig:Fig_APC}.
The \textit{APC} is a 1$\times$1 convolution that has spatially variant learnable kernel weights for each pixel feature. Thus, the \textit{APC} can obtain optimized feature maps by performing linear projection per pixel feature.
The \textit{APC} consists of \textit{N} pointwise convolutions. 
In Figure \ref{fig:Fig_APC}, \textit{EM} denotes element-wise multiplication, also known as the Hadamard product between each output feature maps of the pointwise convolutions and input feature map.
And \textit{ES} denotes element-wise summation for the \textit{N} feature maps.
Note that the adaptive convolution kernel space is determined by output feature maps of \textit{N} pointwise convolutions.

Three loss functions are used to train the model as
$L = L_{1} + L_{DCT} + L_{\nabla}$, 
where $L_{DCT}$ is the DCT loss, $L_{\nabla}$ is the differential content loss. The $L_{DCT}$ and $L_{\nabla}$ can help alleviate over-smoothness in the demoired image and tend to reconstruct image for high frequency components \cite{cheon2018generative}.

\subsection{iipl-hdu team}

\begin{figure}[h]
	\centering
	\includegraphics[width=1.0\columnwidth]{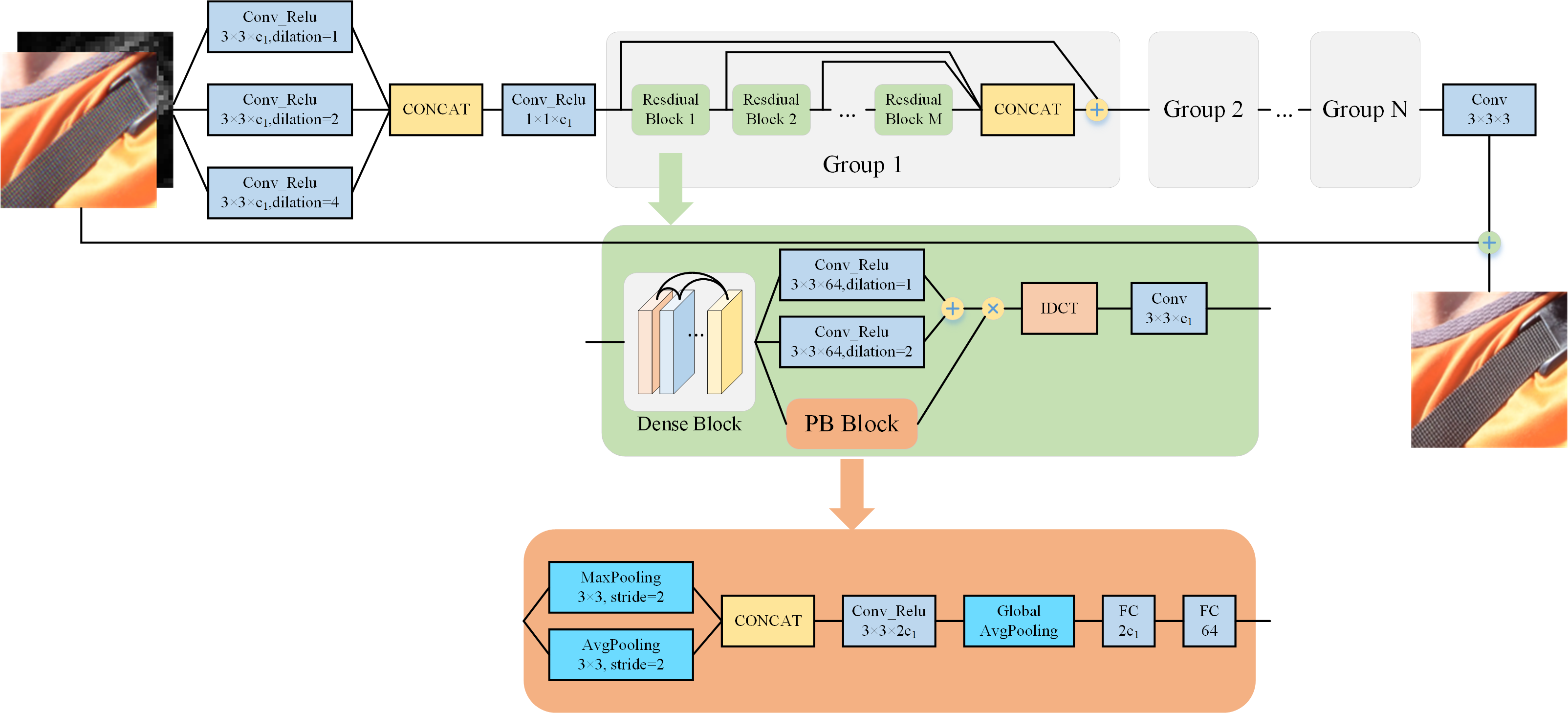}
	\caption{Architecture of the proposed baseline.}
	\label{fig2}
\end{figure}

\begin{figure}[h]
	\centering
		\vspace{-10pt}
	\begin{subfigure}{.22\textwidth}
		\includegraphics[width=\columnwidth]{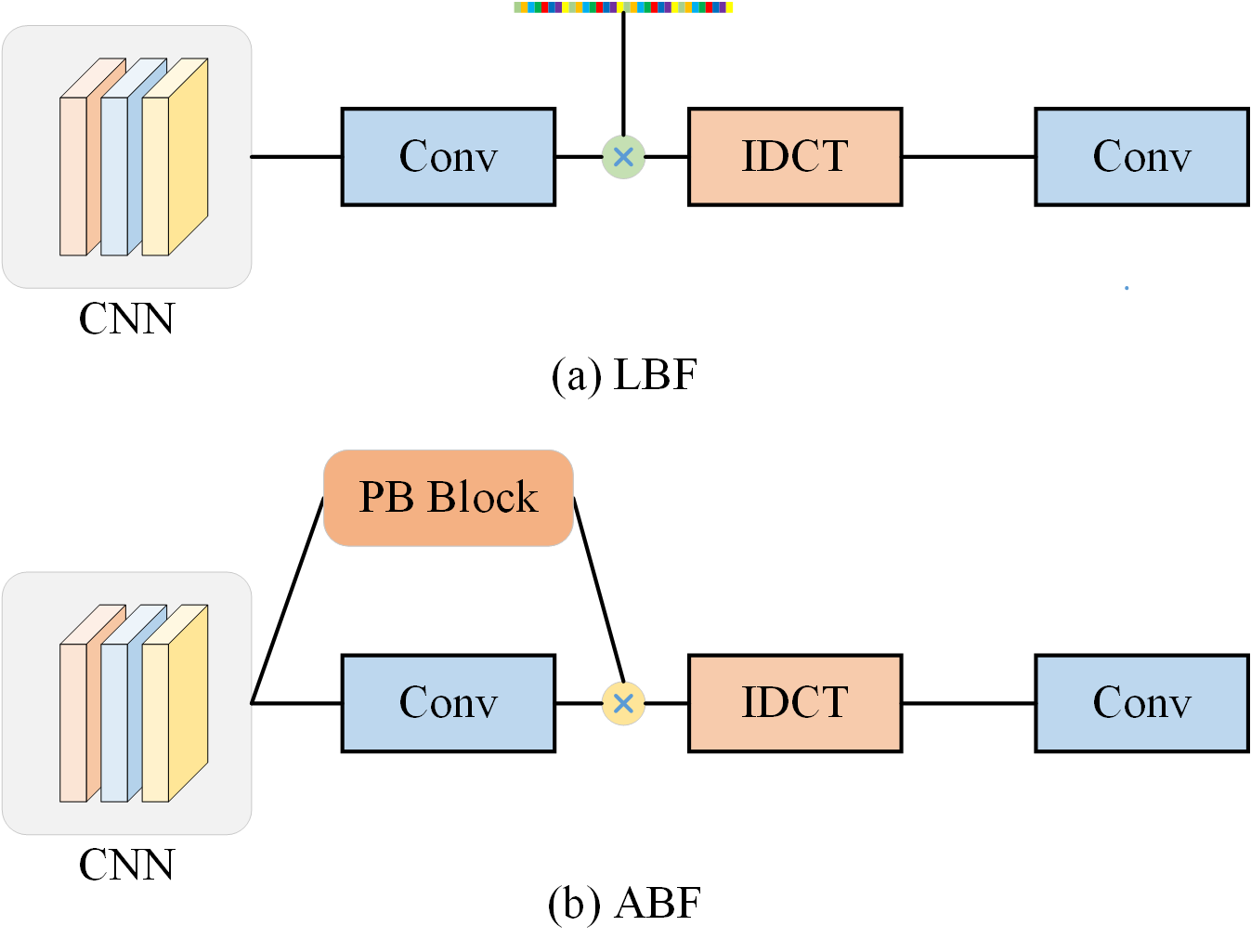}
	\end{subfigure}
	\begin{subfigure}{.24\textwidth}
		\includegraphics[width=0.95\columnwidth]{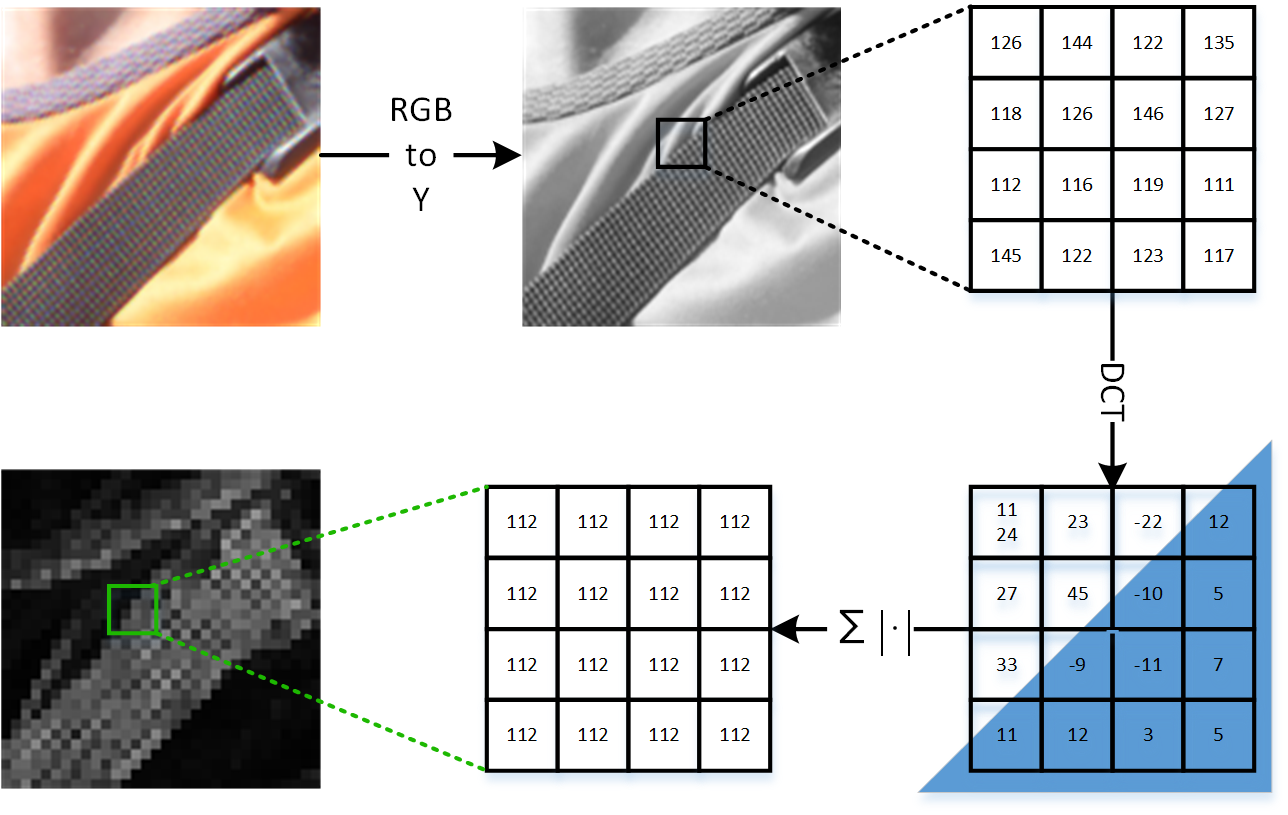}
	\end{subfigure}
	\caption{Proposed modules by \textbf{iipl-hdu} team. Left: the comparison of LBF (a) and ABF (b). Right: the pipeline of producing the label map.}
	\label{fig:proposed_modules}
\end{figure}

The \textbf{iipl-hdu} team proposed a \textbf{CNN-based adaptive bandpass filter for image demoireing}, where Adaptive Bandpass Filte (ABF) is designed. This work is mainly inspired by the team iipl-hdu's previous work, the learnable bandpass filter~\cite{MBCNN,IDCN} (LBF). The new method introduced a predictive bandpass (PB) block to adaptively predict the passband along with the image texture changing. Fig.~\ref{fig:proposed_modules}
shows the difference between ABF and LBF.
ABF was used to construct the residual block, and the architecture of GRDN~\cite{kim2019grdn} was adopted to construct the baseline. The architecture of proposed baseline is shown in Fig.~\ref{fig2}. The baseline is constructed by $N$ groups, and each group contains $M$ residual blocks. In the residual block, let's define the input as a $c_1$-channel feature map, then the dense block includes 5 densely connected dilated convolution layers. The dilation rates of the 5 layers are $1,2,3,2,1$, and all of them output a $c_2$-channel feature map.

\textbf{Label Map:} Realizing that the moire effect only appears in high-frequency areas, a labeling method was proposed to generate the label map, highlighting the high-frequency area in the image. The label map will be concatenated to the moire image and sent to the CNN. The steps shown in Fig.~\ref{fig:proposed_modules} were followed to generate the label map from the moire input.

\subsection{Mac AI team}

\begin{figure}[h]
	\centering
	\includegraphics[width=\linewidth]{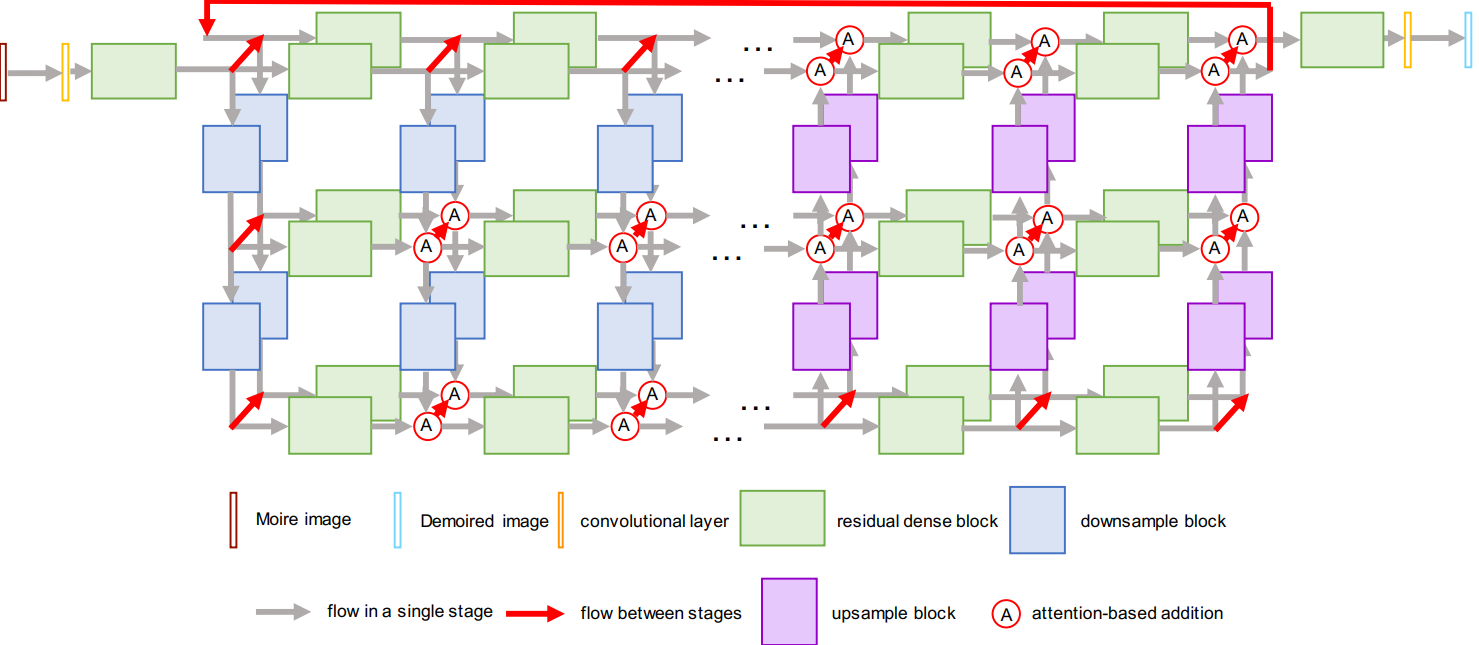}
	\caption{The main architecture of CubeDemoireNet.}
	\label{fig:CubeDemoireNet}
	\vspace{-2mm}
\end{figure}

The \textbf{Mac AI} team proposed \textit{CubeDemoireNet: Enhanced Multi-Scale Network for Image Demoireing}. CubeDemoireNet is inspired by \cite{liu2019griddehazenet}.  It enhances the conventional multi-scale network by 1) facilitating information exchange across different scales and stages to alleviate the underlying bottleneck issue, and 2) employs the attention mechanism to identify the dynamic Moir\'{e} patterns, and thus eliminate them effectively with the preservation of image texture.

Due to the fact that dynamic Moir\'{e} patterns usually embed in a broad range of frequencies, the benefit of a multi-scale structure is evident. However, the conventional multi-scale network has a hierarchical design that often suffers a bottleneck issue caused by insufficient information flow among different scales. The Mac AI team proposed a new network, named CubeDemoireNet, which promotes information exchange across different scales and stages via dense connections using upsampling/downsampling blocks to alleviate the underlying bottleneck issue prevalent in conventional multi-scale networks. Moreover, CubeDemoireNet further integrates the attention mechanism to preserve image texture while eliminating the dynamic Moir\'{e} patterns. Fig.~\ref{fig:CubeDemoireNet} demonstrates the main architecture of the proposed CubeDemoireNet. The network consists of two convolutional layers and three blocks that are residual dense block (RDB), upsampling block, and downsampling block respectively. More specifically, the same settings in \cite{zhang2018residual} are adopted in the proposed RDB. The upsampling and downsampling blocks are comprised of only two convolutional layers to adjust the spatial resolutions accordingly. As for the attention mechanism, the one employed in SENet \cite{hu2018squeeze} is chosen. 

In single and burst demoireing tracks, the proposed CubeDemoireNet is used as the backbone. In CubeDemoireNet, the row, column, and stage is set to 3, 12, and 3 respectively for both tracks. Moreover, it was found that adding a refinement network can further improve the demoireing performance. Therefore, a modified U-net \cite{ronneberger2015u} is adopted for both tracks. Specifically, for the burst demoireing track, to align the burst images in accordance with the reference one, the Pyramid, Cascading and Deformable (PCD) proposed by \cite{wang2019edvr} is employed.

\subsection{MoePhoto team}

\begin{figure}[h]
 \vspace{-5pt}
	\begin{center}
		\includegraphics[width=1.0\linewidth]{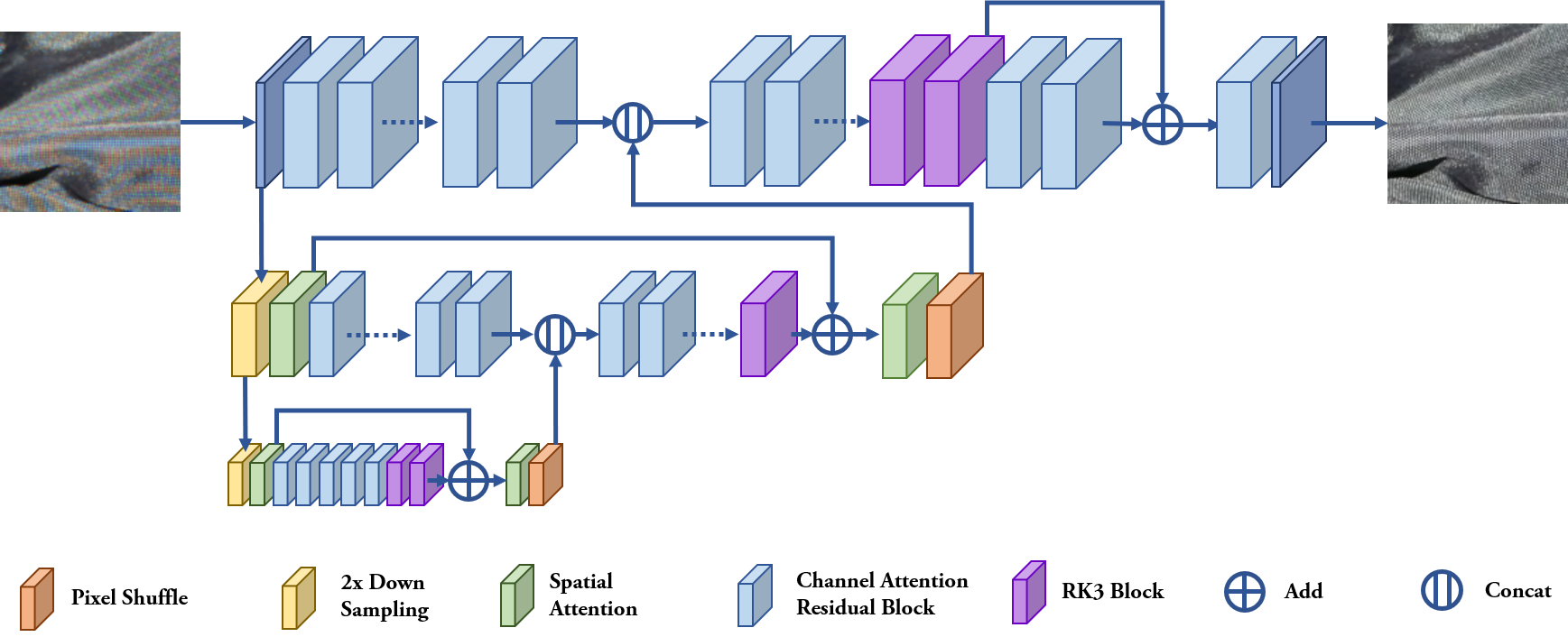}
	\end{center}
	 \vspace{-5pt}
	\caption{Overall structure of MSFAN.}
	\label{fig:short}
	 \vspace{-10pt}
\end{figure}   

The \textbf{MoePhoto} team proposed \textit{Multi-scale feature aggregation network for image demoiréing}. 
This is a multi-scale feature aggregation network for image demoiréing, and dubbed MSFAN, see Figure \ref{fig:short}. Based on \cite{sun2018moire,cheng2019multi}, MSFAN incorporates a multi-branch framework to encode the image with the original resolution features and $2\times$, $4\times$ down sampling features to  obtain different feature representations. In each branch, local residual learning is used to remove moire pattern and recover image details. Then the features are upsampled with sub pixel convolution and concatenated to the upper level for effective feature aggregation. MSFAN uses channel attention and spatial attention to further enhance the information among the channels and grab the non-local spatial information among features with different spatial resolutions. L1 Charbonnier loss and L1 wavelet loss are used to calculate the difference between the output image and the ground truth and update the network parameters.

\subsection{DGU-CILAB}

\begin{figure}[t]
  \centering
  \includegraphics[width=0.45\textwidth]{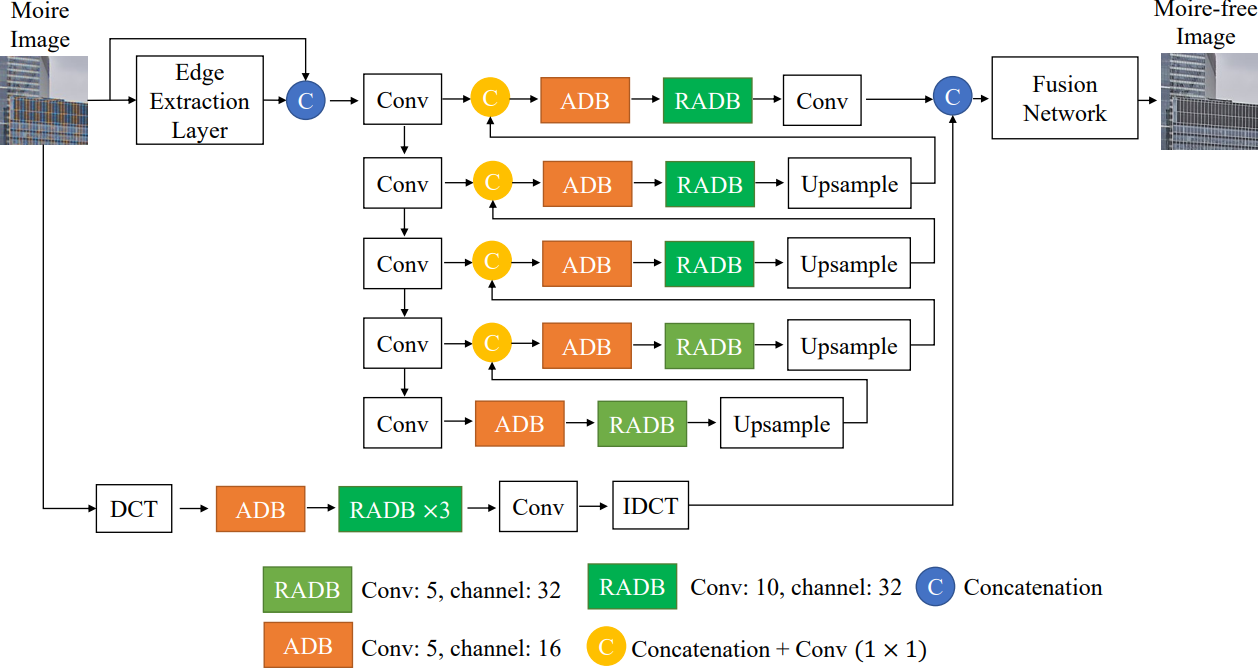}
  \caption{The architecture of the proposed dual-domain demoireing network for Track-1.}
  \label{fig:architecture}
   \vspace{-10pt}
\end{figure}

The \textbf{DGU-CILAB} team proposed \textit{Dual-Domain Deep Convolutional Neural Networks for Demoireing} \cite{CVPRW2020_Vien}, which exploit the complex properties of moiré patterns in multiple domains, {\it i.e.}, pixel domain and frequency domain. In the pixel domain, multi-scale features are employed to remove the moiré artifacts associated with specific frequency bands through multi-resolution feature maps. In the frequency domain, inputs are transformed to spectral sub-bands using the discrete cosine transform (DCT). Then, DGU-CILAB designed a network that processes DCT coefficients to remove moiré artifacts. Next, a dynamic filter generation network \cite{jia2016dynamic} is developed to learn dynamic blending filters. Finally, the results from the pixel domain and frequency domain are combined by blending filters to yield the clean images. The \textbf{Pixel Network} is composed of multiple branches of different resolutions. 
Each branch is a stack of attention dense block (ADB) and residual attention dense block (RADB), which are based on convolutional block attention module (CBAM) \cite{woo2018cbam}, dense block (DB), and residual dense block (RDB) \cite{zhang2018residual}. The \textbf{Frequency Network} processes the DCT coefficients of the input image to remove moiré artifacts in a frequency domain. The final output image is generated by applying the inverse discrete cosine transform (IDCT). The \textbf{Fusion Network} using dynamic filter network \cite{jia2016dynamic} is developed to take a pair of results from the pixel network and frequency network as input and outputs blending filters, which are then used to process the inputs to yield the final moiré-free image. To train the network, first the pixel network and the frequency network are trained separately. Then, after fixing them, the fusion network is trained. For the burst track, information of each image in the sequence is exploited and aligned with the reference image using the attention network.

\subsection{Image Lab team}

\begin{figure}[t]
  \centering
     \includegraphics[width=0.45\textwidth, height=0.3\textwidth]{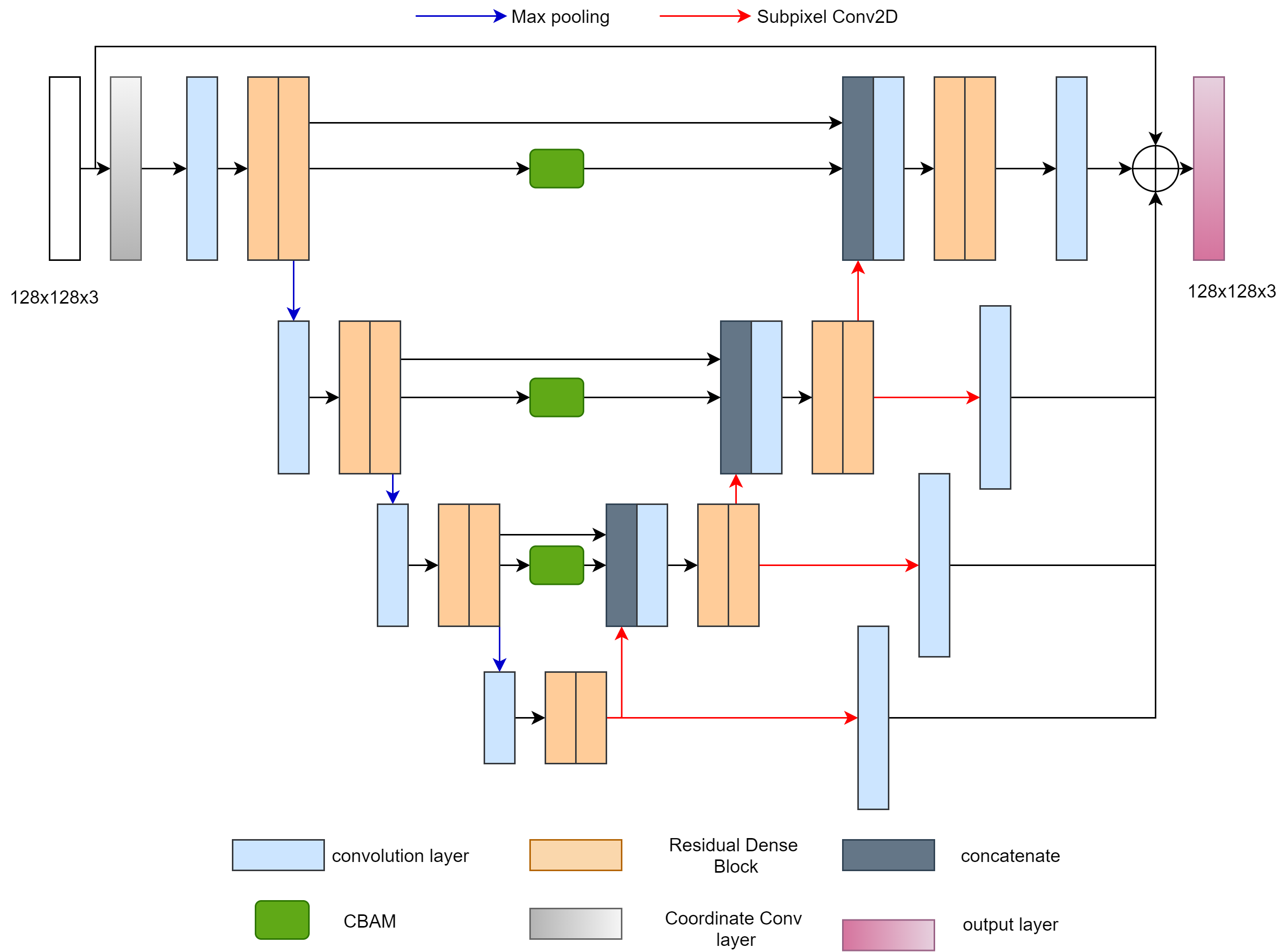}
  \caption{Proposed multilevel hyper vision net architecture.}
  \label{fig:hypervisionnet}
  \vspace{-10pt}
\end{figure}

The \textbf{Image Lab} team proposed \textit{Moire Image Restoration using Multilevel Hyper Vision Net}. 
The proposed architecture\cite{sabarinathan2019morie} is shown in Figure \ref{fig:hypervisionnet}, 
where the input image pixels mapped to the Cartesian coordinate space with the help of a coordinate convolution layer \cite{liu2018intriguing}. The output of coordinate convolution layers passed to encoder block. The convolutional layer and the residual dense attention blocks are utilized for better performance. The proposed network has the properties of an encoder and decoder structure of a vanilla U-Net \cite{sabarinathan2019hyper,ronneberger2015u}. During down-sampling three blocks have been used in the encoder phase. In each block, the first encoder block is a 3$\times$3 convolutional layer, followed by two residual dense attention blocks added and at the next block convolution layer with stride 2 used for down sampling. In the decoder phase, the same blocks have been used except the down sampling layer, which replaced with a super pixel convolutional layer. Output features of the encoder are fed to the convolution block attention module \cite{woo2018cbam} (CBAM) and in the skip connection upsampled features concatenated with encoder block and the output of the CBAM block. Inspired by the Hyper Vision Net \cite{sabarinathan2019hyper} model, in this work the three hyper vision layers in the decoder part were introduced. The output of these hyper vision layers are fused and supervised to obtain the enhanced image. The loss function is $Loss = MSE+(1-SSIM)+SOBEL_{loss}$.

\subsection{VAADER team}

The \textbf{VAADER} team's method is inspired by MBCNN of the AIM2019 challenge \cite{AIM19demoireMethods} and uses a CNN-based multiscale approach. A multi loss extracted from the flow at different scales is used to train the network. A \textit{pre-training} is performed on a mix of AIM2019 challenge patches \cite{AIM19demoireDataset} and this year challenge samples. The
final model is obtained through a second training stage using only the current challenge dataset. Indeed, the past year dataset has several differences including large luminance shifts between input and reference that do not appear in this year's dataset.

\subsection{CET\_CVLab team}

The \textbf{CET\_CVLab} team proposed \textit{Single Image Demoireing With Enhanced Feature Supervised Deep Convolutional Neural Network}.
The architecture follows a CNN based encoder-decoder structure. 
The base network was proposed for image deraining \cite{xiang2019single} and has two convolutional encoders and one decoder i.e, two inputs and one output. 
Encoder-I is a sequence of downsamplers that extract the features of the degraded image at different scales. 
Encoder-II, which has similar structure as encoder-I, extracts the features of the moire-free image at different scales during training. 
The decoder is a sequence of upsamplers that reconstruct the moire-free images from encoder-I's output. 
The features of moire interfered image can be brought closer to that of moire-free image by minimizing their distance of separation in feature space. This is realized by incorporating feature loss during training. Feature loss is the mean of weighted sum of L1 distance between the features of the degraded image and moire-free image at different scales. 
To enhance the network, a block of residual dense network (RDN) \cite{zhang2018residual} with channel attention (CA) \cite{zhang2018image} was introduced before downsampling and upsampling to extract the features in the current scale level. The proposed solution has 6 different scale levels. The model complexity can be adjusted by adding or removing the scale levels

\subsection{sinashish team}

The \textbf{sinashish} team proposed the use of feature fusion attention for the demoireing of moired images inspired by FFA-Net~\cite{qin2019ffa}. The method was trained with an $L2$ loss instead of the $L1$ loss proposed in the FFA-Net paper.

\subsection{NTIREXZ team}

The \textbf{NTIREXZ} team followed the work of FFA-Net \cite{qin2019ffa} to estimate the undesired texture using global residual learning. Attention mechanisms, including both channel attention and pixel-wise attention, are used alongside a multi-scale approach for efficient learning.

\section{Conclusion}

This paper reviews the image demoireing challenge that was part of the NTIRE2020 workshop. 
First, the paper describes the challenge including the new dataset, CFAMoire that was created for participants to develop their demoire methods as well as evaluate proposed solutions. 
The challenge consisted of two tracks (single and burst) that considered different inputs to the demoire approaches submitted by participants. 
The tracks had 142 and 99 registered participants, respectively, and a total of 14 teams competed in the final testing phase. 
The paper summarizes the results of the two tracks, and then describes each of the approaches that competed in the challenge. 
The entries span the current the state-of-the-art in the image demoireing problem. We hope this challenge and its results will inspire additional work in the demoire problem, which is becoming an increasingly important image quality challenge.

\section*{Acknowledgements}
We thank the NTIRE 2020 sponsors: Huawei, OPPO, Voyage81, MediaTek, DisneyResearch$\mid$Studios, and Computer Vision Lab (CVL) ETH Zurich.

\appendix

\section{Teams}
\label{sec:appendix}

\footnotesize{

\textbf{0. NTIRE 2020 team}
\begin{itemize}[noitemsep]
\item \textbf{Members:} \\
Shanxin Yuan$^1$ (\textcolor{cyan}{shanxin.yuan@huawei.com}) \\
Radu Timofte$^2$  (\textcolor{cyan}{radu.timofte@vision.ee.ethz.ch})\\
Ale\v{s} Leonardis$^1$   (\textcolor{cyan}{ales.leonardis@huawei.com})\\
Gregory Slabaugh$^1$  (\textcolor{cyan}{gregory.slabaugh@huawei.com}) 
\item \textbf{Affiliations:} \\
$^1$Huawei Noah's Ark Lab \hspace*{4mm}\\
$^2$ETH Z\"{u}rich, Switzerland
\end{itemize}

\textbf{1. HaiYun team}
\begin{itemize}[noitemsep]
\item \textbf{Members:} 
Xiaotong Luo (\textcolor{cyan}{xiaotluo@qq.com})$^1$, Jiangtao Zhang$^1$, Ming Hong$^1$, Yanyun Qu$^1$, Yuan Xie$^2$, Cuihua Li$^1$
\item \textbf{Affiliations:} \\
$^1$ Xiamen University, China \\
$^2$ East China Normal University, China
\end{itemize}

\textbf{2. OIerM team}
\begin{itemize}[noitemsep]
\item \textbf{Members:} 
Dejia Xu (\textcolor{cyan}{dejia@pku.edu.cn})$^1$, Yihao Chu$^2$, Qingyan Sun$^3$
\item \textbf{Affiliations:} \\
$^1$ Peking University, China\\
$^2$ Beijing University of Posts and Telecommunications, China\\
$^3$ Beijing Jiaotong University, China
\end{itemize}

\textbf{3. Alpha team}
\begin{itemize}[noitemsep]
\item \textbf{Members:} Shuai Liu$^1$ (\textcolor{cyan}{18601200232@163.com}), Ziyao Zong$^1$, Nan Nan$^1$, Chenghua Li$^2$   
\item \textbf{Affiliations:} \\
$^1$North China University of Technology, China\\
$^2$Institute of Automation, Chinese Academy of Sciences, China
\end{itemize}

\textbf{4. Reboot team}
\begin{itemize}[noitemsep]
\item \textbf{Members:} 
Sangmin Kim (\textcolor{cyan}{ksmh1652@gmail.com}), Hyungjoon Nam, Jisu Kim, Jechang Jeong
\item \textbf{Affiliation:} 
Hanyang University, Korea
\end{itemize}

\textbf{5. LIPT team}
\begin{itemize}[noitemsep]
\item \textbf{Members:} 
Manri Cheon (\textcolor{cyan}{manri.cheon@lge.com}), Sung-Jun Yoon, Byungyeon Kang, Junwoo Lee
\item \textbf{Affiliation:} 
LG Electronics
\end{itemize}

\textbf{6. iipl-hdu team}
\begin{itemize}[noitemsep]
\item \textbf{Members:} 
Bolun Zheng (\textcolor{cyan}{zhengbolun1024@163.com})
\item \textbf{Affiliation:}\
Hangzhou Dianzi University, China
\end{itemize}

\textbf{7. Mac AI team}
\begin{itemize}[noitemsep]
\item \textbf{Members:} 
Xiaohong Liu (\textcolor{cyan}{liux173@mcmaster.ca}),  Linhui Dai, Jun Chen 
\item \textbf{Affiliation:} 
McMaster University, Canada
\end{itemize}

\textbf{8. MoePhoto team}
\begin{itemize}[noitemsep]
\item \textbf{Members:} 
Xi Cheng (\textcolor{cyan}{chengx@njust.edu.cn}), Zhenyong Fu, Jian Yang
\item \textbf{Affiliation:} 
Nanjing University of Science and Technology, China
\end{itemize}

\textbf{9. DGU-CILAB team}
\begin{itemize}[noitemsep]
\item \textbf{Members:} 
Chul Lee (\textcolor{cyan}{chullee@dongguk.edu}), An Gia Vien, Hyunkook Park
\item \textbf{Affiliation:}
Dongguk University, South Korea
\end{itemize}

\textbf{10. Image Lab team}
\begin{itemize}[noitemsep]
\item \textbf{Members:} 
Sabari Nathan (\textcolor{cyan}{sabarinathantce@gmail.com})$^1$ , M.Parisa Beham$^2$ ,  S Mohamed Mansoor Roomi$^3$ 
\item \textbf{Affiliations:} \\
$^1$ Couger Inc ,Tokyo,Japan\\
$^2$ Sethu Institute of Technology, India\\
$^3$ Thiagarajar college of Engineering, India
\end{itemize}

\textbf{11. VAADER team}
\begin{itemize}[noitemsep]
\item \textbf{Members:} 
Florian Lemarchand (\textcolor{cyan}{florian.lemarchand@insa-rennes.fr})$^1$, Maxime Pelcat$^1$, Erwan Nogues$^{1,2}$
\item \textbf{Affiliation:} \\
$^1$ Univ. Rennes, INSA Rennes, IETR - UMR CNRS 6164, France\\
$^2$ DGA-MI, Bruz, France
\end{itemize}

\textbf{12. CET\_CVLab team}
\begin{itemize}[noitemsep]
\item \textbf{Members:} 
Densen Puthussery (\textcolor{cyan}{puthusserydensen@gmail.com}), Hrishikesh P S, Jiji C V       
\item \textbf{Affiliation:} 
College of Engineering Trivandrum
\end{itemize}

\textbf{13. Sinashish team}
\begin{itemize}[noitemsep]
\item \textbf{Members:} 
Ashish Sinha (\textcolor{cyan}{asinha@mt.iitr.ac.in})
\item \textbf{Affiliation:} Indian Institute of Technology Roorkee, India
\end{itemize}

\textbf{14. NITREXZ team}
\begin{itemize}[noitemsep]
\item \textbf{Members:} 
Xuan Zhao (\textcolor{cyan}{zxstud@163.com})
\item \textbf{Affiliation:} 
Nanjing University of Aeronautics and Astronautics, China
\end{itemize}

}

{\small
\bibliographystyle{ieee_fullname}
\bibliography{egbib}

\begin{thebibliography}{10}\itemsep=-1pt

\bibitem{abdelhamed2020ntire}
Abdelrahman Abdelhamed, Mahmoud Afifi, Radu Timofte, Michael Brown, et~al.
\newblock Ntire 2020 challenge on real image denoising: Dataset, methods and
  results.
\newblock In {\em CVPRW, 2020}.

\bibitem{abdelhamed2019ntire}
Abdelrahman Abdelhamed, Radu Timofte, and Michael~S. Brown.
\newblock Ntire 2019 challenge on real image denoising: Methods and results.
\newblock In {\em CVPRW, 2019}.

\bibitem{aittala2018burst}
Miika Aittala and Fr{\'e}do Durand.
\newblock Burst image deblurring using permutation invariant convolutional
  neural networks.
\newblock In {\em ECCV, 2018}.

\bibitem{ancuti2018ntire}
Cosmin Ancuti, Codruta~O Ancuti, and Radu Timofte.
\newblock Ntire 2018 challenge on image dehazing: Methods and results.
\newblock In {\em CVPRW, 2018}.

\bibitem{ancuti2020ntire}
Codruta~O. Ancuti, Cosmin Ancuti, Florin-Alexandru Vasluianu, Radu Timofte,
  et~al.
\newblock Ntire 2020 challenge on nonhomogeneous dehazing.
\newblock In {\em CVPRW, 2020}.

\bibitem{arad2020ntire}
Boaz Arad, Radu Timofte, Yi-Tun Lin, Graham Finlayson, Ohad Ben-Shahar, et~al.
\newblock Ntire 2020 challenge on spectral reconstruction from an rgb image.
\newblock In {\em CVPRW, 2020}.

\bibitem{FFCC}
Jonathan Barron and Yun-Ta Tsia.
\newblock Fast fourier color constancy.
\newblock In {\em CVPR, 2017}.

\bibitem{blau20182018}
Yochai Blau, Roey Mechrez, Radu Timofte, Tomer Michaeli, and Lihi Zelnik-Manor.
\newblock The 2018 pirm challenge on perceptual image super-resolution.
\newblock In {\em ECCV, 2018}.

\bibitem{cai2019ntire}
Jianrui Cai, Shuhang Gu, Radu Timofte, and Lei Zhang.
\newblock Ntire 2019 challenge on real image super-resolution: Methods and
  results.
\newblock In {\em CVPRW, 2019}.

\bibitem{gcnet}
Yue Cao, Jiarui Xu, Stephen Lin, Fangyun Wei, and Han Hu.
\newblock Gcnet: Non-local networks meet squeeze-excitation networks and
  beyond.
\newblock {\em arXiv, 2019}.

\bibitem{cheng2019multi}
Xi Cheng, Zhenyong Fu, and Jian Yang.
\newblock Multi-scale dynamic feature encoding network for image
  demoir{\'e}ing.
\newblock {\em arXiv, 2019}.

\bibitem{cheon2018generative}
Manri Cheon, Jun-Hyuk Kim, Jun-Ho Choi, and Jong-Seok Lee.
\newblock Generative adversarial network-based image super-resolution using
  perceptual content losses.
\newblock In {\em ECCV, 2018}.

\bibitem{choi2017deep}
Jae-Seok Choi and Munchurl Kim.
\newblock A deep convolutional neural network with selection units for
  super-resolution.
\newblock In {\em CVPRW, 2017}.

\bibitem{ding2019acnet}
Xiaohan Ding, Yuchen Guo, Guiguang Ding, and Jungong Han.
\newblock Acnet: Strengthening the kernel skeletons for powerful cnn via
  asymmetric convolution blocks.
\newblock In {\em ICCV, 2019}.

\bibitem{fuoli2020ntire}
Dario Fuoli, Zhiwu Huang, Martin Danelljan, Radu Timofte, et~al.
\newblock Ntire 2020 challenge on video quality mapping: Methods and results.
\newblock In {\em CVPRW, 2020}.

\bibitem{gharbi2016deep}
Micha{\"e}l Gharbi, Gaurav Chaurasia, Sylvain Paris, and Fr{\'e}do Durand.
\newblock Deep joint demosaicking and denoising.
\newblock {\em TOG, 2016}.

\bibitem{hdrnet}
Micha{\"e}l Gharbi, Jiawen Chen, Jonathan~T Barron, Samuel~W Hasinoff, and
  Fr{\'e}do Durand.
\newblock Deep bilateral learning for real-time image enhancement.
\newblock {\em TOG, 2017}.

\bibitem{gu2019div8k}
Shuhang Gu, Andreas Lugmayr, Martin Danelljan, Manuel Fritsche, Julien Lamour,
  and Radu Timofte.
\newblock Div8k: Diverse 8k resolution image dataset.
\newblock In {\em ICCVW, 2019}.

\bibitem{gu2019brief}
Shuhang Gu and Radu Timofte.
\newblock A brief review of image denoising algorithms and beyond.
\newblock In {\em CiML, 2019}.

\bibitem{hasinoff2016burst}
Samuel~W Hasinoff, Dillon Sharlet, Ryan Geiss, Andrew Adams, Jonathan~T Barron,
  Florian Kainz, Jiawen Chen, and Marc Levoy.
\newblock Burst photography for high dynamic range and low-light imaging on
  mobile cameras.
\newblock {\em TOG, 2016}.

\bibitem{hu2018squeeze}
Jie Hu, Li Shen, and Gang Sun.
\newblock Squeeze-and-excitation networks.
\newblock In {\em CVPR, 2018}.

\bibitem{Ignatov_2017_ICCV}
Andrey Ignatov, Nikolay Kobyshev, Radu Timofte, Kenneth Vanhoey, and Luc
  Van~Gool.
\newblock Dslr-quality photos on mobile devices with deep convolutional
  networks.
\newblock In {\em ICCV, 2017}.

\bibitem{TGA2020}
Takashi Isobe1, Songjiang Li, Xu Jia, Shanxin Yuan, Gregory Slabaugh, Chunjing
  Xu, Ya-Li Li, Shengjin Wang, and Qi Tian.
\newblock Video super-resolution with temporal group attention.
\newblock In {\em CVPR, 2020}.

\bibitem{jaderberg2015spatial}
Max Jaderberg, Karen Simonyan, Andrew Zisserman, and koray kavukcuoglu.
\newblock Spatial transformer networks.
\newblock In {\em NeurIPS, 2015}.

\bibitem{jia2016dynamic}
Xu Jia, Bert De~Brabandere, Tinne Tuytelaars, and Luc~V Gool.
\newblock Dynamic filter networks.
\newblock In {\em NeurIPS, 2016}.

\bibitem{kim2019grdn}
Dong-Wook Kim, Jae Ryun~Chung, and Seung-Won Jung.
\newblock Grdn: Grouped residual dense network for real image denoising and
  gan-based real-world noise modeling.
\newblock In {\em CVPRW, 2019}.

\bibitem{kim2020c3net}
Sangmin Kim, Hyungjoon Nam, Jisu Kim, and Jechang Jeong.
\newblock C3net: Demoiréing network attentive in channel, color and
  concatenation.
\newblock In {\em CVPRW, 2020}.

\bibitem{liu2018intriguing}
Rosanne Liu, Joel Lehman, Piero Molino, Felipe~Petroski Such, Eric Frank, Alex
  Sergeev, and Jason Yosinski.
\newblock An intriguing failing of convolutional neural networks and the
  coordconv solution.
\newblock In {\em NeurIPS, 2018}.

\bibitem{Liu2020MMDM}
Shuai Liu, Chenghua Li, Nan Nan, Ziyao Zong, and Ruixia Song.
\newblock {MMDM}: Multi-frame and multi-scale for image demoiréing.
\newblock In {\em CVPRW, 2020}.

\bibitem{liu2019griddehazenet}
Xiaohong Liu, Yongrui Ma, Zhihao Shi, and Jun Chen.
\newblock Griddehazenet: Attention-based multi-scale network for image
  dehazing.
\newblock In {\em ICCV, 2019}.

\bibitem{lugmayr2019aim}
Andreas Lugmayr, Martin Danelljan, Radu Timofte, et~al.
\newblock Aim 2019 challenge on real-world image super-resolution: Methods and
  results.
\newblock In {\em ICCVW, 2019}.

\bibitem{lugmayr2020ntire}
Andreas Lugmayr, Martin Danelljan, Radu Timofte, et~al.
\newblock Ntire 2020 challenge on real-world image super-resolution: Methods
  and results.
\newblock In {\em CVPRW, 2020}.

\bibitem{nah2019ntire}
Seungjun Nah, Sungyong Baik, Seokil Hong, Gyeongsik Moon, Sanghyun Son, Radu
  Timofte, and Kyoung Mu~Lee.
\newblock Ntire 2019 challenge on video deblurring and super-resolution:
  Dataset and study.
\newblock In {\em CVPRW, 2019}.

\bibitem{nah2020ntire}
Seungjun Nah, Sanghyun Son, Radu Timofte, Kyoung~Mu Lee, et~al.
\newblock Ntire 2020 challenge on image and video deblurring.
\newblock In {\em CVPRW, 2020}.

\bibitem{niklaus2017video2}
Simon Niklaus, Long Mai, and Feng Liu.
\newblock Video frame interpolation via adaptive separable convolution.
\newblock In {\em ICCV, 2017}.

\bibitem{qin2019ffa}
Xu Qin, Zhilin Wang, Yuanchao Bai, Xiaodong Xie, and Huizhu Jia.
\newblock Ffa-net: Feature fusion attention network for single image dehazing.
\newblock {\em arXiv, 2019}.

\bibitem{romano2016raisr}
Yaniv Romano, John Isidoro, and Peyman Milanfar.
\newblock Raisr: rapid and accurate image super resolution.
\newblock {\em TCI, 2016}.

\bibitem{ronneberger2015u}
Olaf Ronneberger, Philipp Fischer, and Thomas Brox.
\newblock U-net: Convolutional networks for biomedical image segmentation.
\newblock In {\em MICCAI, 2015}.

\bibitem{sabarinathan2019hyper}
D Sabarinathan, M~Parisa Beham, SM Roomi, et~al.
\newblock Hyper vision net: Kidney tumor segmentation using coordinate
  convolutional layer and attention unit.
\newblock {\em arXiv, 2019}.

\bibitem{sabarinathan2019morie}
D Sabarinathan, M~Parisa Beham, SM Roomi, et~al.
\newblock Moire image restoration using multi level hyper vision net, 2020.

\bibitem{shi2016real}
Wenzhe Shi, Jose Caballero, Ferenc Husz{\'a}r, Johannes Totz, Andrew~P Aitken,
  Rob Bishop, Daniel Rueckert, and Zehan Wang.
\newblock Real-time single image and video super-resolution using an efficient
  sub-pixel convolutional neural network.
\newblock In {\em CVPR, 2016}.

\bibitem{su2019pixel}
Hang Su, Varun Jampani, Deqing Sun, Orazio Gallo, Erik Learned-Miller, and Jan
  Kautz.
\newblock Pixel-adaptive convolutional neural networks.
\newblock In {\em CVPR, 2019}.

\bibitem{coral}
Baochen Sun and Kate Saenko.
\newblock Deep coral: Correlation alignment for deep domain adaptation.
\newblock In {\em ECCV, 2016}.

\bibitem{sun2018moire}
Yujing Sun, Yizhou Yu, and Wenping Wang.
\newblock Moir{\'e} photo restoration using multiresolution convolutional
  neural networks.
\newblock {\em TIP, 2018}.

\bibitem{tian2019decoders}
Zhi Tian, Tong He, Chunhua Shen, and Youliang Yan.
\newblock Decoders matter for semantic segmentation: Data-dependent decoding
  enables flexible feature aggregation.
\newblock In {\em CVPR, 2019}.

\bibitem{timofte2017ntire}
Radu Timofte, Eirikur Agustsson, Luc Van~Gool, Ming-Hsuan Yang, and Lei Zhang.
\newblock Ntire 2017 challenge on single image super-resolution: Methods and
  results.
\newblock In {\em CVPRW, 2017}.

\bibitem{Timofte_2018_CVPR_Workshops}
Radu Timofte, Shuhang Gu, Jiqing Wu, Luc Van~Gool, Lei Zhang, Ming-Hsuan Yang,
  Muhammad Haris, et~al.
\newblock Ntire 2018 challenge on single image super-resolution: Methods and
  results.
\newblock In {\em CVPRW, 2018}.

\bibitem{SevenWays}
Radu Timofte, Rasmus Rothe, and Luc Van~Gool.
\newblock Seven ways to improve example-based single image super resolution.
\newblock In {\em CVPR, 2016}.

\bibitem{CVPRW2020_Vien}
An~Gia Vien, Hyunkook Park, and Chul Lee.
\newblock Dual-domain deep convolutional neural networks for image demoireing.
\newblock In {\em CVPRW, 2020}.

\bibitem{wang2019edvr}
Xintao Wang, Kelvin~CK Chan, Ke Yu, Chao Dong, and Chen Change~Loy.
\newblock Edvr: Video restoration with enhanced deformable convolutional
  networks.
\newblock In {\em CVPRW, 2019}.

\bibitem{woo2018cbam}
Sanghyun Woo, Jongchan Park, Joon-Young Lee, and In So~Kweon.
\newblock Cbam: Convolutional block attention module.
\newblock In {\em ECCV, 2018}.

\bibitem{xiang2019single}
Peng Xiang, Lei Wang, Fuxiang Wu, Jun Cheng, and Mengchu Zhou.
\newblock Single-image de-raining with feature-supervised generative
  adversarial network.
\newblock {\em SPL, 2019}.

\bibitem{Luo_2020_CVPR_Workshops}
Luo Xiaotong, Zhang Jiangtao, Hong Ming, Qu Yanyun, Xie Yuan, and Li Cuihua.
\newblock Deep wavelet network with domain adaptation for single image
  demoireing.
\newblock In {\em CVPRW, 2020}.

\bibitem{OIerM_2020_Demoire}
Dejia Xu, Yihao Chu, and Qingyan Sun.
\newblock Moir{\'e} pattern removal via attentive fractal network.
\newblock In {\em CVPRW, 2020}.

\bibitem{yang2020derain}
Wenhan Yang, Shiqi Wang, Dejia Xu, Xiaodong Wang, and Jiaying Liu.
\newblock Towards scale-free rain streak removal via self-supervised fractal
  band learning.
\newblock In {\em AAAI, 2020}.

\bibitem{yu2019local}
Songhyun Yu and Jechang Jeong.
\newblock Local excitation network for restoring a jpeg-compressed image.
\newblock {\em IEEE Access, 2019}.

\bibitem{yu2019deep}
Songhyun Yu, Bumjun Park, and Jechang Jeong.
\newblock Deep iterative down-up cnn for image denoising.
\newblock In {\em CVPRW, 2019}.

\bibitem{AIM19demoireDataset}
Shanxin Yuan, Radu Timofte, Gregory Slabaugh, and Ales Leonardis.
\newblock Aim 2019 challenge on image demoireing: dataset and study.
\newblock In {\em ICCVW, 2019}.

\bibitem{AIM19demoireMethods}
Shanxin Yuan, Radu Timofte, Gregory Slabaugh, Ales Leonardis, et~al.
\newblock Aim 2019 challenge on image demoireing: methods and results.
\newblock In {\em ICCVW, 2019}.

\bibitem{zhang2020ntire}
Kai Zhang, Shuhang Gu, Radu Timofte, et~al.
\newblock Ntire 2020 challenge on perceptual extreme super-resolution: Methods
  and results.
\newblock In {\em CVPRW, 2020}.

\bibitem{DnCNN}
Kai Zhang, Wangmeng Zuo, Yunjin Chen, Deyu Meng, and Lei Zhang.
\newblock Beyond a gaussian denoiser: Residual learning of deep cnn for image
  denoising.
\newblock {\em TIP, 2017}.

\bibitem{zhang2018image}
Yulun Zhang, Kunpeng Li, Kai Li, Lichen Wang, Bineng Zhong, and Yun Fu.
\newblock Image super-resolution using very deep residual channel attention
  networks.
\newblock In {\em ECCV, 2018}.

\bibitem{zhang2019residual}
Yulun Zhang, Kunpeng Li, Kai Li, Bineng Zhong, and Yun Fu.
\newblock Residual non-local attention networks for image restoration.
\newblock {\em arXiv, 2019}.

\bibitem{zhang2018residual}
Yulun Zhang, Yapeng Tian, Yu Kong, Bineng Zhong, and Yun Fu.
\newblock Residual dense network for image super-resolution.
\newblock In {\em CVPR, 2018}.

\bibitem{IDCN}
Bolun Zheng, Yaowu Chen, Xiang Tian, Fan Zhou, and Xuesong Liu.
\newblock Implicit dual-domain convolutional network for robust color image
  compression artifact reduction.
\newblock {\em TCSVT, 2019}.

\bibitem{MBCNN}
Bolun Zheng, Shanxin Yuan, Gregory Slabaugh, and Ales Leonardis.
\newblock Image demoireing with learnable bandpass filters.
\newblock In {\em CVPR, 2020}.

\end{thebibliography}
}

\end{document}